\documentclass[preprint,12pt]{elsarticle}

\usepackage{amssymb}
\usepackage{amsfonts}       
\usepackage{nicefrac}       
\usepackage{microtype}      
\usepackage{bm}
\usepackage{subfig}
\usepackage{graphicx}
\usepackage{amsfonts}
\usepackage{mathrsfs}
\usepackage{amsmath}
\usepackage{amssymb}
\usepackage{amsthm}
\usepackage{caption3}
\usepackage{algorithm}
\usepackage{algorithmic}
\usepackage{multirow}	
\usepackage{indentfirst}
\usepackage{booktabs}
\usepackage{array}
\usepackage{pdfpages}
\usepackage{float}
\usepackage{epstopdf}
\usepackage{epsfig}
\newtheorem{theorem}{Theorem}

\journal{Neural Networks}

\begin{document}

\begin{frontmatter}



\title{Low Rank Regularization: A review}


\address[label2]{School of Computer Science, Northwestern Polytechnical University, Xi'an, 710072, Shaanxi, P.R. China.}
\address[label3]{School of Cybersecurity, Northwestern Polytechnical University, Xi'an, 710072, Shaanxi, P.R. China.}
\address[label1]{Center for OPTical IMagery Analysis and Learning (OPTIMAL), Northwestern Polytechnical University, Xi'an, 710072, Shaanxi, P.R. China.}
\author[label2,label1]{Zhanxuan Hu}
\author[label2,label1]{Feiping Nie}
\author[label3,label1]{Rong Wang*}
\cortext[mycorrespondingauthor]{Corresponding author: Rong Wang~(wangrong07@tsinghua.org.cn)}
\author[label2,label1]{Xuelong Li}

\address{}

\begin{abstract}
Low Rank Regularization~(LRR), in essence, involves introducing a low rank or approximately low rank assumption to target we aim to learn, which has achieved great success in many data analysis tasks. Over the last decade, much progress has been made in theories and applications. Nevertheless, the intersection between these two lines is rare. In order to construct a bridge between practical applications and theoretical studies, in this paper we provide a comprehensive survey for LRR. Specifically, we first review the recent advances in two issues that all LRR models are faced with: $(1)$ rank-norm relaxation, which seeks to find a relaxation to replace the rank minimization problem; $(2)$ model optimization, which seeks to use an efficient optimization algorithm to solve the relaxed LRR models. For the first issue, we provide a detailed summarization for various relaxation functions and conclude that the non-convex relaxations can alleviate the punishment bias problem compared with the convex relaxations. For the second issue, we summary the representative optimization algorithms used in previous studies, and analysis their advantages and disadvantages. As the main goal of this paper is to promote the application of non-convex relaxations, we conduct extensive experiments to compare different relaxation functions. The experimental results demonstrate that the non-convex relaxations generally provide a large advantage over the convex relaxations. Such a result is inspiring for further improving the performance of existing LRR models.
\end{abstract}

\begin{keyword}


low rank, regularization, optimization
\end{keyword}

\end{frontmatter}


\section{Introduction}\label{sec:introduction}
In many fields the data we analysis are generally a set of vectors, matrices or tensors. Some examples include voices in signal processing, documents in natural language processing, users' records in recommender systems, images/videos in computer vision, and DNA microarrays in bioinformatics. Such data are in general high dimensional, which brings a series of challenges to subsequent data analysis tasks. Fortunately, the  high-dimensional data generally have some specific structural characteristics, e.g., sparsity. Compressed sensing and sparse representation are two powerful tools to analysis the order-one signals data with sparsity, both of them naturally fit for vectors and achieve tremendous success in several applications. However, we are often faced with matrices or tensors data, such as images, users' records and genetic microarrays. Then we are naturally faced with a problem: how to leverage the  sparsity of matrices or tensors? Low rank is a powerful tool to this issue, which is a metric to second order~(i.e., matrix) sparsity~\cite{Lin2017A}.

\begin{figure*}[t]
	\centering
\includegraphics[width=0.5\linewidth]{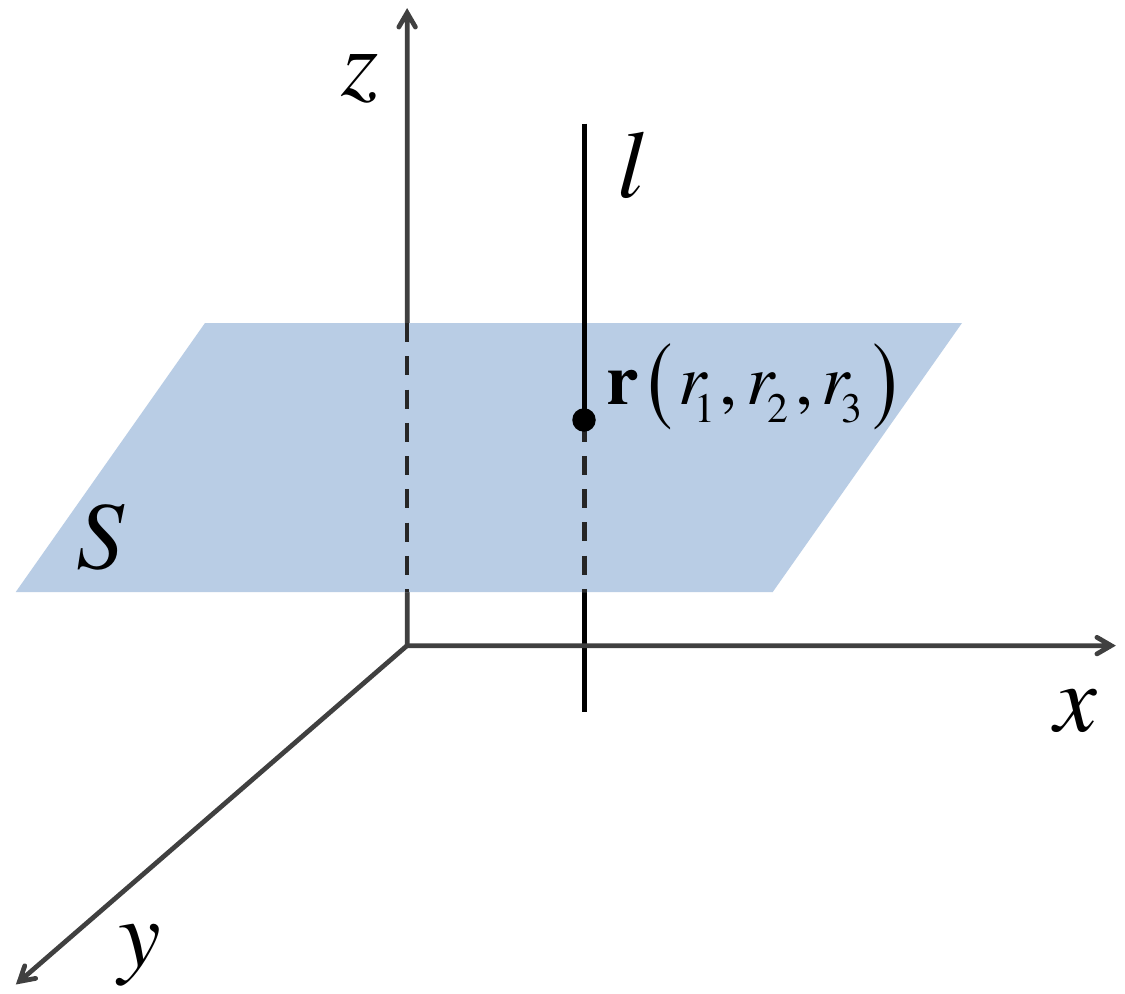}
	\caption{An illustration to matrix completion using low rank regularition.}
	\label{Fig:1s}
\end{figure*}

A typical example is recommendation system, where we have an uncompleted rating matrix and aim to leverage the known rates of users on some items to infer their ratings on others. Such a problem refers to matrix completion in machine learning.
 Mathematically, we provide a simple example in Figure~\ref{Fig:1s}, where $\mathbf{r}=(r_1,r_2,r_3)$ is a column vector~(or row vector) of rating matrix $\mathbf{R}$. The location of $\mathbf{r}$ in $3$D space can be exactly confirmed when the values of $(r_1,r_2,r_3)$ are known. Nevertheless, the only thing we can  determine is that $\mathbf{r}$ locates in the
line $l$ when $r_3$ is absent. Fortunately, the location of 3D vector $\mathbf{r}$~(i.e., the value of $r_3$) is inferable when it locates in a $2$D subspace $\mathcal{S}$. Hence, a natural assumption for matrix completion task is that all columns or rows locate in a common low-dimensional subspace. That is the rating matrix is intrinsically low rank. Such an assumption is natural and reasonable, because users ratings for some items generally influenced by ratings for other kinds of items. Without the Low Rank Regularization~(LRR), it becomes impossible to infer new users ratings on items. 

Over the last decade, LRR sparked a large research interest from various machine learning models including matrix completion. And such models have achieved great success in many fields such as computer vision and data mining. Specifically, recent advances in LRR can be roughly divided into two groups: \emph{theoretical studies} and \emph{practical applications}.

\textbf{Theoretical studies.} The main lines of research are rank relaxation and optimization algorithms. Optimizing the LRR models inevitably involves solving a rank minimization problem, which is known to be NP-hard. An alternative is to relax
the rank-norm~\footnote{Note that rank-norm is not valid norm. Here, "rank-norm" is for convenience.} using nuclear norm. Such that the relaxed problem is convex and can be readily solved by existing convex optimization tools. Nevertheless, nuclear norm treats all singular values equally, leading to a
bias to the matrix with small singular values. Towards this issue, arguably one of the most popular methods is using non-convex relaxations such as  TNN~(Truncated Nuclear Norm)~\cite{DBLP:journals/pami/HuZYLH13}, WNN~(Weighted Nuclear Norm)~\cite{DBLP:conf/cvpr/GuZZF14,DBLP:journals/ijcv/GuXMZFZ17,DBLP:conf/aaai/ZhongXLLC15} and Schatten-p norm~\cite{DBLP:journals/jmlr/MohanF12,DBLP:conf/aaai/NieHD12,shang2017bilinear,DBLP:conf/aaai/ShangLC16,DBLP:journals/tip/LuLY15}.  Although non-convex relaxations are often helpful in reducing the \emph{relaxation gap}, they bring a new challenging problem to algorithm designers. This is because traditional optimization algorithms are  tailored to convex problems and inapplicable in non-convex problems. To this end, in recent years, a class of non-convex optimization algorithms have been developed, such as IRW~(Iteratively Re-weighted method)~\cite{candes2008enhancing,ochs2015iteratively,daubechies2010iteratively}, ADMM~(Alternating Direction Method of Multipliers)~\cite{lu2018unified}, APG~(Accelerated Proximal Gradient) method~\cite{li2015accelerated,ghadimi2016accelerated,DBLP:journals/corr/abs-1708-00146,yao2016efficient}, and Frank-Wolfe Algorithm~\cite{zhang2012accelerated}.

\textbf{Practical applications.} LRR is a popular tool in a wide range of data mining and computer vision tasks. Examples include, but not limited to, recommendation systems~\cite{xu2020multi,li2019both}, Visual Tracking~\cite{DBLP:journals/ijcv/SuiTZW18}, $3$D Reconstruction~\cite{agudo2018image,DBLP:journals/ijcv/DaiLH14,DBLP:conf/cvpr/ZhuHTL14}, Salient Object Detection~\cite{DBLP:journals/pami/PengLLHXM17,DBLP:conf/cvpr/ShenW12,DBLP:conf/bmvc/ZouKLR13,gao2014block}.
 The models used in these tasks are generally derived from some basic machine learning models including Matrix Completion ~\cite{DBLP:journals/tit/CandesT10,DBLP:conf/ijcai/NieHH17}, Subspace Clustering~\cite{DBLP:journals/pami/LiuLYSYM13} and Multi-Task Learning~\cite{DBLP:conf/aaai/HanZ16,DBLP:journals/pami/ZhenYHL18}.

In order to promote the interaction between  \emph{Theoretical studies} and \emph{Practical applications}, some excellent reviews and surveys over LRR~\cite{DBLP:journals/csur/ZhouYZY14,ma2018efficient,lin2017low,Lin2017A} have been made. However, most of them specialized in single group, e.g., low-rank matrix learning and its applications in image analysis~\cite{DBLP:journals/csur/ZhouYZY14,lin2017low,Lin2017A}, optimization algorithms used in low rank matrix learning~\cite{ma2018efficient}. Besides, these studies often revolve around the nuclear norm regularization and ignore the non-convex relaxations. But the superiority of non-convex regularizers over nuclear norm has been verified in many theoretical studies~\cite{DBLP:journals/jmlr/MohanF12,DBLP:conf/aaai/NieHD12,shang2017bilinear,DBLP:conf/aaai/ShangLC16,DBLP:journals/tip/LuLY15,DBLP:journals/pami/HuZYLH13}. To this end, in this paper we provide a new review for LRR and pay more attention to the non-convex relaxations and corresponding optimization methods. The main contributions of this paper are summarized as follows:
\begin{itemize}
  \item We first provide an exhaustive analysis for commonly used relaxations, including convex and non-convex relaxations. Then we summarize some representative optimization algorithms for solving the relaxed low rank models. Both of them are absent in previous reviews and surveys.
  \item We conduct a great many of experiments to compare the performance of different relaxations. The experimental results demonstrate that the non-convex relaxations generally provide a large advantage over the convex
relaxations. Such a result is useful for promoting the application of non-convex relaxations in solving practical issues.
\end{itemize}
It is worth emphasizing that this work mainly focuses on the LRR models rather than low rank matrix learning, and the former can be considered as a subproblem of the latter. For low rank matrix learning, we recommend the book~\cite{lin2017low}. The rest of this paper is organized as follows. We first review the \emph{Theoretical studies} over LRR in Sect.~\ref{sec:2}. Then, in Sect~\ref{sec:3} we summarize the machine learning models using LRR and their applications in solving practical issues. To further analysis the relaxation functions, we conduct numerous experiments and report the results in Sect~\ref{sec:5}. Finally,  we conclude this paper in Sect.~\ref{sec:6} and discuss the future work over LRR.

\textbf{Notations.} In this work, the vectors and matrices are denoted by lowercase boldface letters $\mathbf{x}$ and uppercase boldface letters $\mathbf{X}$ respectively. $\mathbf{x}^i$, $\mathbf{x}_i$ and $\mathbf{X}_{ij}$ denote the $i$th row, $j$th column, and $(i,j)$th element of matrix $\mathbf{X}$. $Tr(\mathbf{X})$ denotes the trace of matrix $\mathbf{X}$. We summarize the vectors/matrix norms in Table~\ref{tab:norm}, where $\sigma_i(\mathbf{X})$ is the $i$th singular value of matrix  $\mathbf{X}$, and function $f(x)$ is a  function over $\sigma_i(\mathbf{X})$.

\begin{table}[]
\centering
\caption{Vevtor/matrix Norms}\label{tab:norm}
\begin{tabular}{|c|l|l|}
\hline
Type                    & norm    & Defination \\ \hline\hline
\multirow{3}{*}{Vector norm} & $\ell_0$-norm      &   $\|\mathbf{x}\|_{0} = \sum_{i=1}^{d}|x_i|^0$         \\ \cline{2-3}
                        & $\ell_1$-norm       &  $\|\mathbf{x}\|_{1} = \sum_{i=1}^{d}|x_i|$          \\ \cline{2-3}
                        & $\ell_2$-norm       &  $\|\mathbf{x}\|_{2} = (\sum_{i=1}^{d}|x_i|^2)^{\frac{1}{2}}$           \\ \hline\hline
\multirow{6}{*}{Matrix norm} & Rank-norm     &   $\|\mathbf{X}\|_{r} = \sum_{i=1}^{k}\sigma_i(\mathbf{X})^0$         \\ \cline{2-3}
                        & Nuclear-norm  &   $\|\mathbf{X}\|_{*} = \sum_{i=1}^{k}\sigma_i(\mathbf{X})$         \\ \cline{2-3}
                        & $f$-rank-norm       &    $\|\mathbf{X}\|^f_{*} = \sum_{i=1}^{k}f(\sigma_i(\mathbf{X}))$        \\ \cline{2-3}
                        & Frobenius-norm        &   $\|\mathbf{X}\|_F = (Tr(\mathbf{X}\mathbf{X}^T))^\frac{1}{2}$  \\ \cline{2-3}
                        & $\ell_{2,1}$-norm       &    $\|\mathbf{X}\|_{2,1}= \sum_{i=1}^{n}\|\mathbf{x}^i\|_2$   \\ \cline{2-3}
                        & $\ell_{1}$-norm       &    $\|\mathbf{X}\|_{1}= \sum_{i,j}|X_{ij}|$   \\ \cline{2-3}
                         \hline
\end{tabular}
\end{table}

\section{Theoretical studies over Low Rank Regulaization}\label{sec:2}
\subsection{Problem formulation}
Over the last decade, LRR models have attracted much attention due to its success in various fields. All these models assume that the targets to be learnt
lie near single or multiple low-dimensional subspaces, and use low-rankness as a regularizer to build the model. Here, we provide a formal formulation:
\begin{equation}\label{eq:1}
\min_{\mathbf{X} }\ \ \mathcal{L}(\mathbf{ X})+ \lambda \|\mathbf{X}\|_r \ \ s.t. \ \ \mathbf{X} \in \mathcal{C}\,,
\end{equation}
where $\mathcal{L}(\mathbf{X})$ represents the loss term that depends on the tasks we deal with, $\|\mathbf{X}\|_r$~(the rank of $\mathbf{X}$) represents the regularization term, while $\lambda$ is a parameter balancing these two terms. Besides, $\mathcal{C}$ represents the constraints over $\mathbf{X}$.  We are often faced with two issues in solving the LRR models. First, \textbf{relaxation to LRR}. Matrix rank minimization problem is known to be NP-hard, hence we need to find an alternative to replace the rank-norm $\|\mathbf{X}\|_r$. Second, \textbf{optimization algorithm to LRR}, the relaxed models is often non-convex/non-smooth, and we need to find an efficient optimization algorithm to solve the relaxed models. For these two lines of research are generally suitable for most existing LRR models, we group them into \emph{Theoretical studies}. Next, we review the recent advances in each of them.
\subsection{Relaxations to LRR} \label{sec:2.1}
Given a matrix $\mathbf{X}$ and its ordered singular values $(\sigma_1(\mathbf{X}),\sigma_2(\mathbf{X}),\ldots,\sigma_k(\mathbf{X}))$. The matrix rank-norm is equivalent to $\|\bm{\sigma}\|_0$, where $\bm{\sigma}$ is a vector, and ${\sigma_i} = \sigma_i(\mathbf{X})$.
 Similar to $\ell_0$-norm minimization, matrix rank minimization is known to be NP-hard. An alternative is using a relaxation to replace rank-norm regularization. We denote the relaxed regularizer by
$  \mathcal{R}(\mathbf{X}) = \sum_{i=1}^{k}f(\sigma_i(\mathbf{X}))$,
where $f(x)$ represents a relaxation function. According to the property of $f(x)$, we divide commonly used regularizers into two groups: \emph{convex relaxations} and \emph{non-convex relaxations}. A summarization for these relaxations can be found in Table~\ref{TAB:REGULARIZER}.
\begin{table}[]
\small
\centering
\caption{Commonly used relaxations in LRR. $\gamma$ or $p$ refers to the parameter used in relaxation function, and $\lambda$ represents the regularization parameter. }
\label{TAB:REGULARIZER}
\begin{tabular}{|c|c|}
\hline
   Name           &  $\lambda \mathcal{R} (\mathbf{X})$  \\ \hline\hline
   Nuclear norm   &  $\sum_{i=1}^{k}\lambda\sigma_i$    \\ \hline
                         Elastic-Net~\cite{DBLP:conf/cvpr/KimLO15}   &   $\sum_{i=1}^{k}\lambda(\sigma_i+\gamma\sigma_i^2)$ \\ \hline \hline
  Sp-norm~\citep{DBLP:journals/jmlr/MohanF12,DBLP:conf/aaai/NieHD12}   &  $\sum_{i=1}^{k}\lambda\sigma_i^p$   \\ \hline

                             TNN~\cite{DBLP:journals/pami/HuZYLH13}   &  $\sum_{i=r+1}^{k}\lambda\sigma_i$    \\ \hline
                             PSN~\cite{oh2013partial}&  $\sum_{i=r+1}^{k}\lambda\sigma_i$ \\ \hline
                             WNN~\cite{DBLP:conf/cvpr/GuZZF14,DBLP:journals/ijcv/GuXMZFZ17,DBLP:conf/aaai/ZhongXLLC15} &  $\sum_{i=r+1}^{k}\lambda w_i\sigma_i$ \\ \hline
                             CNN~\cite{DBLP:conf/kdd/SunXY13,DBLP:conf/ijcai/NieHH17}&  $\sum_{i=r+1}^{k}\lambda min(\sigma_i,\theta)$  \\ \hline
                             Capped Sp~\cite{DBLP:conf/ijcai/NieHH17} &  $ \sum_{i=r+1}^{k}\lambda min(\sigma_i^p,\theta)$($0<p<1$)  \\ \hline
                             $\gamma$-Nuclear norm~\cite{DBLP:conf/icdm/KangPC15} &  $ \sum_{i=1}^{k}\frac{\lambda(1+\gamma)\sigma_i}{\gamma+\sigma_i}$  \\ \hline
                             LNN~\cite{DBLP:conf/kdd/PengKLC15} &  $ \sum_{i=1}^{k}\lambda log(\sigma_i+1)$ \\ \hline
                             Logarithm~\cite{friedman2012fast,DBLP:conf/cvpr/LuTYL14}     & $\sum_{i=1}^{k}\frac{\lambda log(\gamma \sigma_i+1)}{log(\gamma+1)}$  \\ \hline
                             ETP~\cite{gao2011feasible,DBLP:conf/cvpr/LuTYL14}    &  $ \sum_{i=1}^{k}\frac{\lambda(1-exp(-\gamma \sigma_i))}{1-exp(-\gamma)}$ \\ \hline
                             Geman~\cite{geman1995nonlinear,DBLP:conf/cvpr/LuTYL14}  &  $ \sum_{i=1}^{k}\frac{\lambda\sigma_i}{\sigma_i+\gamma}$  \\ \hline
                             Laplace~\cite{trzasko2009highly,DBLP:conf/cvpr/LuTYL14} &  $ \sum_{i=1}^{k}\lambda(1-exp(-\frac{\sigma_i}{\gamma}))$ \\ \hline
                             MCP~\cite{zhang2010nearly,DBLP:conf/cvpr/LuTYL14} &  $ \sum_{i=1}^{k} \left\{ \begin{array}{l}
\lambda \sigma_i  - \frac{\sigma_i }{{2\gamma }},\quad if\;\sigma_i  < \gamma \lambda ,\\
\frac{{\gamma {\lambda ^2}}}{2},\quad \;\quad \;\;if\;\sigma_i  \ge \gamma \lambda .
\end{array} \right.$ \\ \hline
                             SCAD~\cite{fan2001variable,DBLP:conf/cvpr/LuTYL14} &  $ \sum_{i=1}^{k} \left\{ \begin{array}{l}
\lambda x,\quad \quad \quad\quad\quad\quad \sigma_i \leq \lambda \\
\frac{{ - {\sigma_i^2} + 2\gamma \lambda \sigma_i - {\lambda ^2}}}{{2\left( {\gamma - 1} \right)}},\quad \lambda  < \sigma_i \leq \lambda \gamma \\
\frac{{{\lambda ^2}\left( {\gamma  + 1} \right)}}{2},\quad \quad \quad \quad \sigma_i > \lambda \gamma
\end{array} \right.$ \\ \hline
\end{tabular}
\end{table}

\subsubsection{Convex relaxations}
Nuclear norm corresponds to the function $f(x)=x$. The connection between it and rank-norm is introduced by~\cite{fazel2001rank}, where the author report a significant Theorem.
\begin{theorem}
  The convex envelope of the function $\phi(\mathbf{X})=\|\mathbf{X}\|_r$, on constraint $\mathcal{C}=\{\mathbf{X}\in \mathbb{R}^{m\times n}|\|\mathbf{X}\|\leq 1\}$, is $\phi_env(\mathbf{X})=\|\mathbf{X}\|_*$. Here, $\|\mathbf{X}\| = \sigma_1(\mathbf{X})$ denotes the spectral norm, i.e., the largest singular value of matrix $\mathbf{X}$.
\end{theorem}
This Theorem demonstrates that we can obtain a lower bound on the optimal solution of the rank minimization problem via solving a heuristic problem~(nuclear norm relaxation).

Although nuclear norm is convex and can be readily solved by existing convex
optimization tools, it often leads to a bias to the matrix with small singular values.
 Recently, a particular convex regularizer was developed in~\cite{DBLP:journals/ijcv/LarssonO16}, which aims to find a convex approximation for $\|\mathbf{X}\|_r+\lambda\|\mathbf{X} -\mathbf{X}_0\|_F^2$ rather than the rank-norm. Hence, this regularizer will be ignored in the rest of this paper. In addition, a Elastic-Net Regularization of Singular Values~(ERSV) has been proposed in~\cite{DBLP:conf/cvpr/KimLO15}, which corresponds to function $f(x)=x+\gamma x^2$.
\subsubsection{Non-convex relaxations}
Although nuclear norm has achieved success in several practical applications, it suffers a well-documented shortcoming that all singular values are simultaneously minimized.
However, in real data larger singular values generally quantify the main information we want to preserve.
To this end, in recent years, a class of non-convex relaxation approaches has permeated the fields of machine learning.

The advantages of non-convex relaxations over nuclear norm are first shown in~\cite{DBLP:journals/jmlr/MohanF12,DBLP:conf/aaai/NieHD12} for dealing with the matrix completion problems. In particular, both of them generalize the nuclear norm to Schatten-$p$ norm~\footnote{Actually, Schatten-$p$ norm refers to the Schatten-$p$ quasi norm when $0<p<1$.}.
Considering that larger singular values should not be punished, Hu~\cite{DBLP:journals/pami/HuZYLH13} propose a Truncated Nuclear Norm~(TNN) which punishes  only the $n-r$ smallest singular values.
A similar regularizer, namely Partial Sum Nuclear Norm~(PSNN), is developed in~\cite{oh2013partial}. Indeed, both TNN and PSNN can be considered as special cases of Capped Nuclear Norm~(CNN) used in~\cite{DBLP:conf/kdd/SunXY13}.
To alleviate rather than abandon the punishment on larger singular values, Gu et al.~\cite{DBLP:conf/cvpr/GuZZF14} propose a weighted nuclear norm:
\begin{equation}\label{eq:tnn}
\textstyle
  \|\mathbf{X}\|_{\mathbf{w}*} = \sum_{i=1}^{k}w_i\sigma_i(\mathbf{X})\,,
\end{equation}
where $\mathbf{w} = [w_1,w_2,\ldots,w_k]$ is weight vector, and $w_i\geq0$ is a weight value assigned to $\sigma_i(\mathbf{X})$.
Obviously, the punishment bias between larger values and small values can be alleviated by assigning small weights to the former and larger weights to the latter. In addition, TNN, PSNN and CNN can be considered as the special cases of WNN with weight vector:
\begin{equation*}
  \mathbf{w} = [\underbrace{0,0,\ldots,0}_{r},\underbrace{1,\ldots,1}_{k-r}]\,.
\end{equation*}

In addition to the relaxations mentioned above, numerous non-convex relaxations derived from sparse learning have been proposed, such as $\gamma$-nuclear norm~\cite{DBLP:conf/icdm/KangPC15}, Log Nuclear Norm~(LNN)~\cite{DBLP:conf/kdd/PengKLC15}, ETP~\cite{gao2011feasible}, Logarithm~\cite{friedman2012fast}, Geman~\cite{geman1995nonlinear}, Laplace~\cite{trzasko2009highly}, MCP~\cite{zhang2010nearly}, and so on. The details of using them to tackle the matrix completion problems can be found in~\cite{DBLP:conf/cvpr/LuTYL14,DBLP:journals/jmlr/YaoK17,DBLP:journals/corr/abs-1708-00146,DBLP:conf/aaai/LuZXYL15}.

Matrix factorization is another method for low-rank regularization, which represents the expected low rank matrix $\mathbf{X}$ with rank $r$ as $\mathbf{X} = \mathbf{U}\mathbf{V}^T$, where $\mathbf{U}\in \mathbb{R}^{m\times r}$ and $\mathbf{V}\in \mathbb{R}^{n\times r}$. Moreover, the equation~\eqref{eq:nuclear1}~\cite{cabral2013unifying} has been used to solve the matrix completion problem.
\begin{equation}\label{eq:nuclear1}
  \|\mathbf{X}\|_* = \min_{\mathbf{A},\mathbf{B}:\mathbf{A}\mathbf{B}^T=\mathbf{X}}\frac{1}{2}(\|\mathbf{A}\|_F^2+\|\mathbf{B}\|_F^2),
\end{equation}
where $\mathbf{A}\in \mathbb{R}^{m\times d}$, $\mathbf{B}\in \mathbb{R}^{n\times d}$ and $d\geq r$. Recently, Shang et al.~\cite{shang2017bilinear} develop two variants of Eq.~\eqref{eq:nuclear1}, namely Double Nuclear norm penalty~($\|\mathbf{X}\|_{D-N}$) and Frobenius/nuclear hybrid norm penalty ($\|\mathbf{X}\|_{F-N}$). Both of them focus on the connection between Schatten-$p$ norm and matrix factorization.
\begin{figure}[t]
	\centering
	\subfloat[Log]{\includegraphics[width=0.33\linewidth]{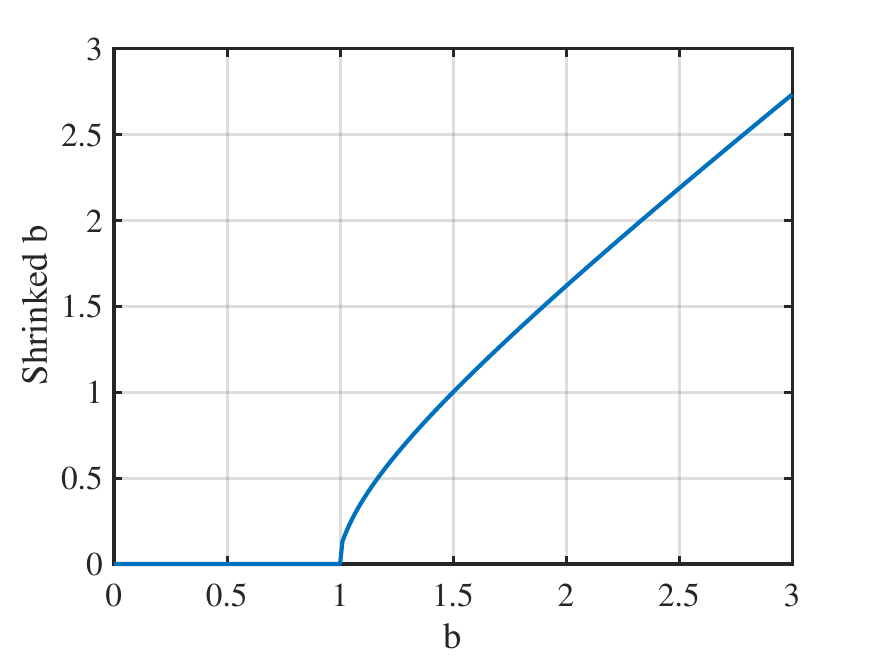}
		\label{Fig1a}}
   \subfloat[ETP]{\includegraphics[width=0.33\linewidth]{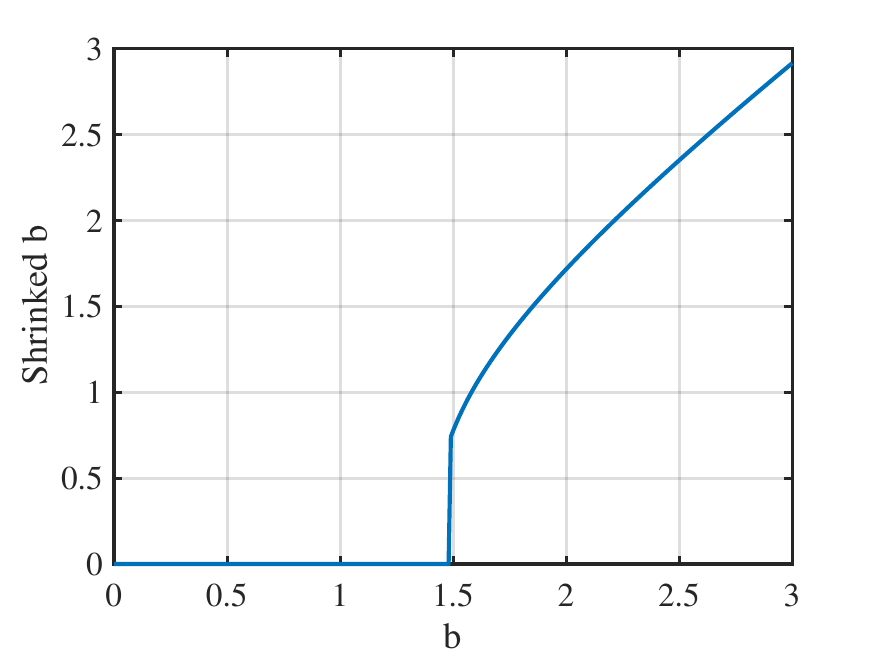}
		\label{Fig1b}}
	\subfloat[Laplace]{\includegraphics[width=0.33\linewidth]{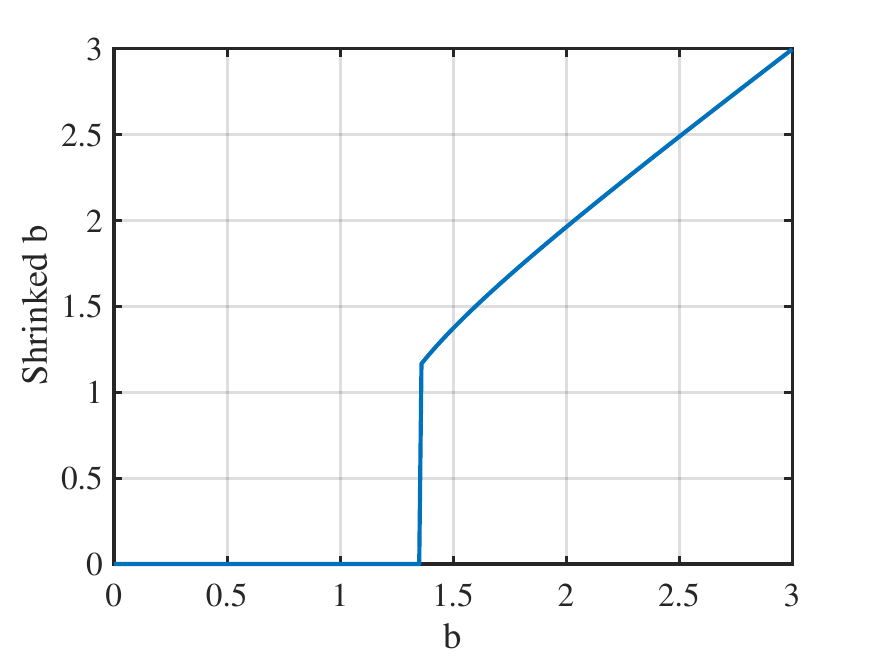}
		\label{Fia1c}}
    \hfil
   \subfloat[$S_p$ with $p=0.1$]{\includegraphics[width=0.33\linewidth]{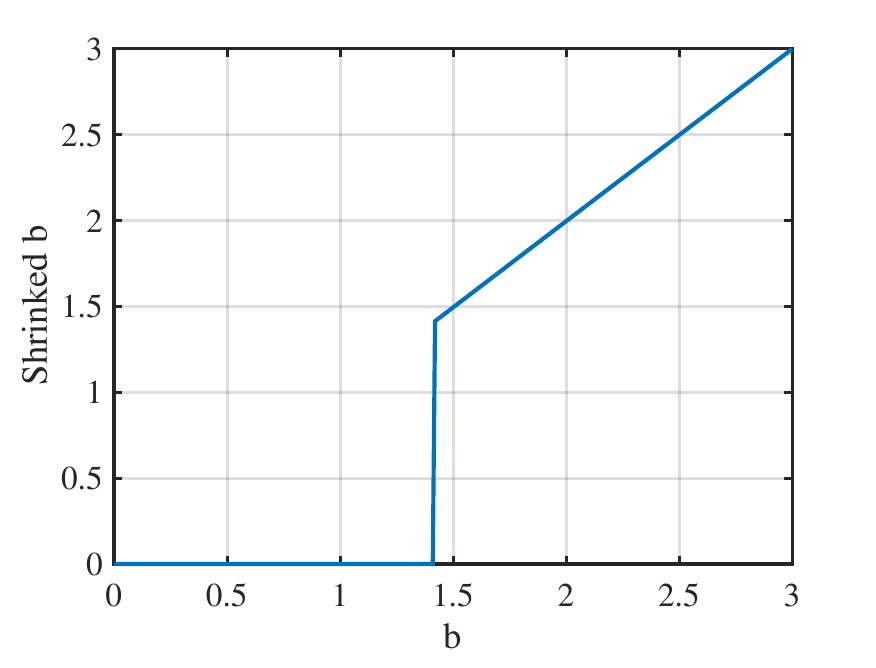}%
		\label{Fig1d}}
   \subfloat[$S_p$ with $p=0.5$]{\includegraphics[width=0.33\linewidth]{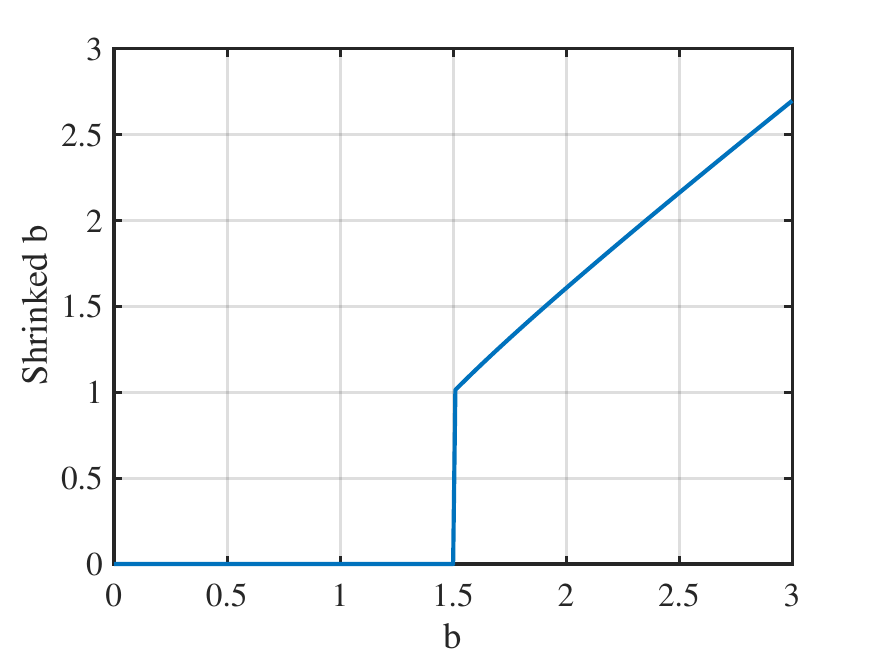}%
		\label{Fig1e}}
	\subfloat[nuclear norm]{\includegraphics[width=0.33\linewidth]{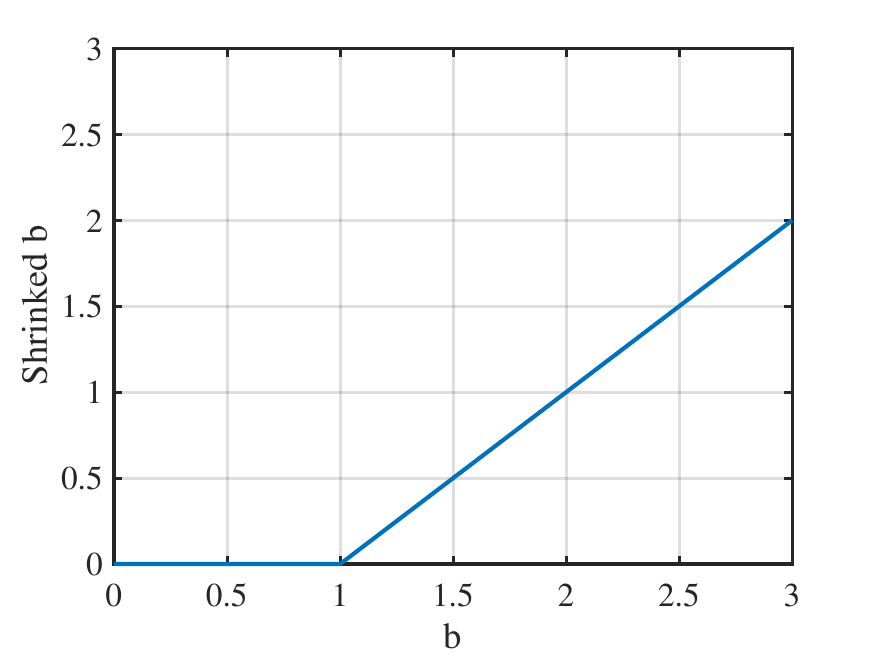}%
		\label{Fia1f}}
	\caption{Comparisons between $b$ and $\rho_f(b,\lambda)$ with $\lambda =1$, where $f$ represents the relaxation function that we select to relax the rank-norm. For ETP, we set $\gamma=1$. For Laplace, we set $\gamma=0.5$.}
	\label{Fig:1}
\end{figure}

\textbf{Closed-form solutions to relaxations}. An interesting property of most existing relaxations mentioned above is that the closed-form solutions can be obtained directly in a specific minimization problem.
\begin{equation}\label{eq:4}
   \min_{\mathbf{X}}\frac{1}{2}\|\mathbf{X}-\mathbf{M}\|_F^2+\lambda \mathcal{R}(\mathbf{X})\,.
\end{equation}
For nuclear norm, i.e., $\mathcal{R}(\mathbf{X})=\|\mathbf{X}\|_*$,  the close-form solution of problem~\eqref{eq:4} can be obtained directly via a Singular Value Thresholding~(SVT) operator $\rho(x,\lambda)$~\cite{DBLP:journals/siamjo/CaiCS10}.
\begin{equation}\label{eq:snn}
\textstyle
  \hat{\mathbf{X}} = \mathbf{U}\rho(\mathbf{S},\lambda)\mathbf{V}^T\,,
\end{equation}
where $\mathbf{U}\mathbf{S}\mathbf{V}^T=\mathbf{M}$ is the SVD of $\mathbf{M}$, and
\begin{equation}\label{eq:snn2}
\rho(\mathbf{S},\lambda)_{ii} = max(S_{ii}-\lambda,0)\,.
\end{equation}
The Eq.~\eqref{eq:snn2} shows that nuclear norm treats all singular values equally and shrinks them with the same threshold $\lambda$. Such a result is consistent with the analysis mentioned above. Besides, Gu et al.~\cite{DBLP:conf/cvpr/GuZZF14} propose a weighted SVT~(WSVT) operator $\rho_{\mathbf {w}}(x,\lambda)$ to weighted nuclear norm, i.e., $\mathcal{R}(\mathbf{X})=\|\mathbf{X}\|_{\mathbf{w}*}$. Correspondingly, we have:
\begin{equation}\label{stw}
\rho_{\mathbf {w}}(\mathbf{S},\lambda)_{ii} = max(S_{ii}-w_i\lambda,0)\,.
\end{equation}
It alleviates the punishment bias between larger singular values
and small singular values via assigning small weights to the former
and larger weights to the latter. Recently, Lu et al.~\cite{DBLP:conf/aaai/LuZXYL15} generalize SVT to a more general case:
\begin{equation}\label{eq:snuclear}
 \min_{\mathbf {X}}\frac{1}{2}\|\mathbf {X}-\mathbf {M}\|_F^2+\lambda\sum_{i=1}^{k}f(\sigma_i(\mathbf {X}))\,.
\end{equation}
where $f(x)$ can be anyone continuous function satisfying the condition: \emph{Function $f(x)$ is concave, nondecreasing and differentiable  on $[o,+\infty)$. Besides, $f(0)=0$ and $\nabla f$ is convex.} Some representatives including Schatten-p norm, MCP~\cite{zhang2010nearly}, Geman~\cite{geman1995nonlinear} and Laplace penalties~\cite{trzasko2009highly} are reported in Table~\ref{TAB:REGULARIZER}. Further, they provide a Generalized Singular Value Thresholding (GSVT) operator $\rho_g(x,\lambda)$. Specifically,
\begin{equation}\label{eq:as}
  \rho_f(x,\lambda) = arg \min_{x\geq0} f_b(x) = \lambda f(x) + \frac{1}{2}(x-b)^2\,,
\end{equation}
where $b$ denotes a singular value of $\mathbf{M}$.
A general solver for finding the optimal solution of the problem~\eqref{eq:as} has been provided in~\cite{DBLP:conf/aaai/LuZXYL15}~\footnote{We provide the codes in: https://github.com/sudalvxin/2018-GSVT.git}. To demonstrate the difference between different relaxations including nuclear norm, we report the shrinkage results returned by them in Figure~\ref{Fig:1}, where the difference between $b$ and $\rho_f(b,\lambda)$ represents the shrink. Figure~\ref{Fig:1} shows that when $b$ takes a small value the shrinkage effect of different relaxations are similar. Nevertheless, when $b$ takes a large value the difference between non-convex relaxations and nuclear norm are significant. Furthermore, the shrink of non-convex relaxations on larger values are very small, which is contrast with the nuclear norm that takes serious shrinks on larger singular values. Hence, using non-convex relaxations can preserve the main information of $\mathbf{M}$. In addition, one can find that non-convex relaxations prefer to generate $0$ singular values, i.e., the low rank solution when regularization parameter $\lambda$ takes the same value.

\subsection{Optimization algorithms to LRR}
Solving a problem with low rank regularization has drawn significant attention, and a great many of specialized optimization approaches have
been proposed.  In this section we review some popular optimization methods that are suitable for both convex relaxations and non-convex relaxations. Note that some traditional optimization methods to LRR models are omitted due to their limitations in solving practical issues. Examples include Singular Value Projection~(SVP) algorithm~\cite{jain2017non} and  fixed-point continuation algorithm~\cite{goldfarb2011convergence}.
\subsubsection{Frank-Wolfe~(FW) Algorithm}
The FW algorithm~\cite{frank1956algorithm,jaggi2013revisiting} is one of the earliest first-order approaches for solving the problems of the form:
\begin{equation}\label{eq:fw}
  \min_{{x}\in \mathcal{C}} f({x}),
\end{equation}
where $x$ can be a vector or matrix, $f({x})$ is Lipschitz-smooth and convex. FW is an iterative method, and at $t+1$ iteration, it updates ${x}$ by
\begin{gather}\label{eq:fw}
   m_t = \arg \min_{m\in\mathcal{C}} m^T\Delta f(x_t), \\
   \gamma_t = \arg \min_{\gamma\in[0,1]} f((1-\gamma)x_t+\gamma m_t)\,,\\
   x_{t+1} = (1-\gamma_t)x_t+\gamma_t m_t\,,
\end{gather}
where Eq.~$(11)$ is a tractable subproblem. The convergence rate of FW algorithm is $\mathcal{O}(1/T)$~\cite{jaggi2013revisiting}. The details of using FW to solve the LRR models with nuclear norm can be found in~\cite{jaggi2013revisiting}. In addition, an improved method for matrix completion problem can be found in~\cite{freund2017extended}. Recently, Yao et al.~\cite{DBLP:journals/jmlr/YaoK17} generalize FW to tackle with non-convex relaxations via redistributing non-convexity of regularization term to loss.

\subsubsection{Proximal Gradient~(PG) algorithm}
Proximal Gradient~(PG) algorithm is commonly used method for solving the unconstraint optimization problem of the form:
\begin{equation}\label{eq:pg}
  \min_{{x}} f({x})+g(x)\,,
\end{equation}
where $f(x)$ is convex and L-Lipschitz smooth, and $g(x)$ is convex. In $t+1$ iteration, PG updates ${x}$ by:
\begin{equation}\label{eq:pg}
  \begin{split}
        x_{t+1} &= \arg \min_{x} f(x_t) + (x-x_t)^T\Delta f(x_t)+\frac{L}{2}\|x-x_t\|_2^2+g(x) \\
       & = \arg \min_{x} \frac{1}{2}\|x-x_t+\frac{1}{L}\Delta f(x_t)\|_2^2 + g(x)\,.
  \end{split}
\end{equation}
The convergence rate of PG is also $\mathcal{O}({1}/{T})$. But, it can be accelerated to $\mathcal{O}({1}/{T^2})$ by incorporating Nesterov's technique~\cite{nesterov1983method}. Ye et al.~\cite{DBLP:conf/icml/JiY09} use APG to solve the nuclear norm minimization problem. The case that $f(x)$ or $g(x)$ is non-convex has been studied in~\cite{li2015accelerated,ghadimi2016accelerated,DBLP:journals/corr/abs-1708-00146,yao2016efficient}. In~\cite{li2015accelerated}, the authors develope two APG-type algorithms, named monotone APG~(mAPG) and non-monotone APG~(nmAPG), respectively. Both of them replace the descent condition used in~\cite{beck2009fast} by a sufficient descent condition. In addition, an Inexact Proximal Gradient algorithm~(IPG) was developed in~\cite{yao2016efficient} to reduce the computation cost caused by two proximal mappings, and a fast proximal algorithm was developed in~\cite{DBLP:journals/corr/abs-1708-00146} to reduce the computation cost caused by conducting SVT over a large scale matrix.
Note that most of existing PG-type algorithms are constructed for solving the problem that is unconstrained~(over desired matrix) and has only one variable to optimize~\footnote{In~\cite{DBLP:journals/corr/abs-1708-00146}, the authors extend it to cope with the problem involved two separable parameter blocks.}.
Hence, although it has sound theoretical guarantee in terms of convergence, it is rarely used in solving the practical issues where various constraints must be considered and multiple variables must be optimized simultaneously. Conversely, generalizing APG methods to tackle with complicated models is a promising direction.
\subsubsection{Iteratively Re-Weighted~(IRW) Algorithm}

The Iteratively Re-Weighted algorithm~(IRW) is primitively designed for solving a class of sparse learning problems of the form~\eqref{eq:fw}~\cite{candes2008enhancing,ochs2015iteratively,daubechies2010iteratively}. It is intrinsically derived from the Majorization-Minimization~(MM) algorithm~\cite{ochs2015iteratively}. Specifically, at $t+1$-th iteration, IRW updates $x$ by:
\begin{equation}\label{eq:pg}
        x_{t+1} = \arg \min_{x} \hat{f}(x,x_t) + g(x)\,,
\end{equation}
where $\hat{f}(x,x_t)$ is a  convex envelope of $f(x)$ at point $x_t$.
The essence of MM is iteratively solving a series of subproblems that are amenable to existing first-order methods. Iterative Reweighted Least Squares algorithms~(IRLS)~\cite{DBLP:journals/jmlr/MohanF12} is a seminal work using IRW to handle the rank minimization problem, and a similar algorithm was developed in~\cite{DBLP:conf/aaai/NieHD12}. Both of them focus on solving the matrix completion problems with Schatten-p norm~($0<p\leq 1$). Recently, Kummerle et al~\cite{kummerle2018harmonic} provide a  Harmonic Mean Iteratively Re-Weighted~(HM-IRW) for such models, which exhibits a locally superlinear rate of convergence under some specific conditions.
 Nie et al.~\cite{nie2018calibrated}  study a more general case that the relaxation function $f(x)$ used in $\mathcal{R}(\mathbf{X})$ is any differentiable concave function.
In addition, an Iteratively Reweighted Nuclear Norm~(IRNN) method has been proposed in~\cite{DBLP:conf/cvpr/LuTYL14}, which can be considered as a combination of PG and IRW. Similar to PG-type method, IRW is constructed for dealing with the problems with single variable. So, its feasibility in practical issues is also limited.

\subsubsection{Alternating Direction Method of Multipliers}
Alternating Direction Method of Multipliers~(ADMM)~\cite{boyd2011distributed} is a popularly used method for solving constraint optimization problem of the form:
\begin{equation}\label{eq:admm}
\begin{split}
    &   \min_{x,y} f(x)+g(y) \\
     & s.t. \ \ \mathbf{A}x+\mathbf{B}y = c\,,
\end{split}
\end{equation}
where $f(x)$ and $g(x)$ are convex. Suppose the augmented Lagrangian function of Eq.~\eqref{eq:admm} is:
\begin{equation}\label{eq:admm2}
  \mathcal{L}(x,y,u) = f(x) +g(y) + u^T(\mathbf{A}x+\mathbf{B}y-c)+\frac{\mu}{2}\|\mathbf{A}x+\mathbf{B}y-c\|_2^2\,,
\end{equation}
where $u$ denotes the Lagrangian multipliers, while $\mu$ denotes the penalty parameter. At $t+1$-th iteration, ADMM update $x$ and other variables by:
\begin{gather}\label{eq:admm3}
  x_{t+1} = \arg \min_{x} \mathcal{L}(x,y_t,u_t),\\
 y_{t+1} = \arg \min_{y} \mathcal{L}(x_t,y,u_t), \\
 u_{t+1} = u_t + \mu (\mathbf{A}x+\mathbf{B}y-c).
\end{gather}
A comprehensive survey and an useful tool for ADMM can be found in~\cite{lu2018unified}, where a unified optimization framework is provided. Theoretically, ADMM converges to a critical point for convex problems. However, its convergence is still an open issue when optimization problem is non-convex. Besides, a practical variant of ADMM, namely Relaxed ADMM, was discussed in~\cite{fang2015generalized,xu2017adaptive}.

\subsubsection{Discussion and summarization}
We have described four representative algorithms for solving the LRR models. A discussion over these algorithms is as follows:
\begin{itemize}
  \item All algorithms can deal with the LRR models with non-convex relaxations. And the core is solving a subproblem iteratively.
  \item FW and PG-type algorithms can deal with the problems with single variable or multiple separable variables. IRW is parameter-free, but it is suitable for the problems
with single variable. For LRR models with multiple variables, IRW can ensure the objective function value of original problem being monotonously decreasing when we use the alternating minimization method to update the variables in subproblem.
  \item  Compared with FW, PG-type and IRW, ADMM is more suitable for complicated problems with equality constraints and multiple non-separable variables. The reason is that in ADMM the low rank regularization
over any one variable can be transferred to an auxiliary variable, and the corresponding subproblems can be solved directly using SVT or GSVT. However, there are two shortcomings in ADMM: parameter selection~($\mu$) and  low convergence speed.
\item The time consuming of IRW is high on large scale data, because it needs to compute the full SVD of $\mathbf{X}$ or the eigenvalue decomposition of $\mathbf{X}\mathbf{X}^T$~($\mathbf{X}^T\mathbf{X}$)~\cite{nie2018calibrated} in subproblem.  For ADMM and PG-type algorithms, the subproblem often has a form of Eq.~\eqref{eq:snuclear}. It is obvious that in Eq.~\eqref{eq:snuclear} the $i$th singular value of $\mathbf{X}$ is zero when $\sigma_i(\mathbf{M})$ is smaller than a specific threshold. That is, only several leading singular values of $\mathbf{M}$ are needed in SVT and GSVT~\cite{DBLP:journals/corr/abs-1708-00146}. Such that the computational complexity of SVT and GSVT can be reduced from $\mathcal{O}(mn^2)$ to $\mathcal{O}(mnr)$ by using PROPACK~\cite{larsen1998lanczos}, where $r$ represents the number of leading singular values. In addition, an useful tool for further reducing the time consuming is using approximate SVT or GSVT ~\cite{DBLP:journals/corr/abs-1708-00146,oh2018fast}.
\end{itemize}

\section{Applications of Low Rank Regularization}\label{sec:3}
Over the last decade, LRR sparked a large
research interest from various machine learning models. According to the \emph{type} of target to be learnt, we roughly group these models into two categories: $(1)$ \emph{data learning models}, the target to be learnt is a data matrix, such as Robust Principal Component Analysis and Robust Matrix Completion; $(2)$ \emph{coefficient learning models}, the target to be learnt is a coefficient~(weight) matrix, such as Subspace Clustering and Multi-Task Learning.  In this section, we first review these four representative machine learning models and then show their applications in practical issues.

\subsection{Applications of LRR in machine learning}\label{sec:3.1}

\subsubsection{Robust Principal Component Analysis~(RPCA)} In most cases, the fundamental assumption for using LRR is that the data we collected lies near some low-dimensional subspaces. For instance, users' records~(such as ratings for movies) in recommender systems, and images in computer vision. In the real world, however, the data are generally corrupted by noise and outliers. RPCA~\citep{candes2011robust} is one of the most significant tools for recovering a low-rank matrix robustly from noisy observations.  Mathematically, RPCA assumes that the data matrix $\mathbf{M}$ is the sum of a low rank matrix $\mathbf{X}$ and a noise matrix $\mathbf{E}$ and use the following LRR models to recover $\mathbf{X}$.

\begin{equation}\label{eq:rpca}
        \min_{\mathbf{X},\mathbf{E}}\ \ \|\mathbf{E}\|_1 + \lambda \|\mathbf{X}\|_r\quad\quad s.t. \ \ \mathbf{M}= \mathbf{X}+\mathbf{E} \,, \\
\end{equation}
where $\|\mathbf{E}\|_1$ is suitable for the sparse noise and can be replaced by other matrix norms.

\subsubsection{Robust Matrix Completion~(RMC)}
RMC~\cite{DBLP:journals/pami/HuZYLH13} is one of the most important variants of RPCA, which considers a general case that some entries of the input data matrix $\mathbf{M}$ are missing, and the known entries are corrupted by noise. The goal of RMC is utilizing the known information to estimate the values of missing entries. The basic assumption used by RMC is that the complete matrix $\mathbf{X}$ we aim to recover is low rank or approximately low rank. Correspondingly, the problem of RMC
can be solved by using a model of the form:
\begin{equation}\label{eq:rmc}
        \min_{\mathbf{X}}\ \ \|\mathcal{P}_\Omega(\mathbf{X}-\mathbf{M})\|_1 + \lambda \|\mathbf{X}\|_r \,, \\
\end{equation}
where $\mathcal{P}_\Omega$ represents a projecting operator, $\Omega$ represents a set recording the indices of known entries. The entries of matrix $\mathcal{P}_\Omega(\mathbf{X})$ are consistent with $\mathbf{X}$ on $\Omega$ and are $0$ on residuals.

\subsubsection{Multi-Task Learning~(MTL)}
Both RPCA and RMC are based on an assumption that the data matrix we are faced with is low rank or approximately low rank. That is, the data we collected is relevant. The relatedness among \emph{different samples} further inspires researchers to explore the relatedness among \emph{different tasks}. Given $K$ relevant tasks $\{\mathcal{T}_i\}_{i=1}^K$ accompanied by feature matrix $\{\mathbf{X}_i\in \mathbb{R}^{n_i\times d}\}_{i=1}^K$ and target vectors $\{\mathbf{y}_i\in \mathbb{R}^{n_i\times 1}\}_{i=1}^K$, we can learn them simultaneously to improve the generalization performance of each one. Such a problem refers to Multi-Task Learning~(MTL). A general LRR model for MTL is:
\begin{equation}\label{eq:mtl}
  \min_{\mathbf{W}} \sum_{i=1}^{K}\|\mathbf{X}_i\mathbf{w}_i-\mathbf{y}_i\|_2^2+\lambda \|\mathbf{W}\|_r\,,
\end{equation}
where $\mathbf{W}\in \mathbb{R}^{d\times K}$ is a weight matrix, and its $i$th column $\mathbf{w}_i$ is the weight vector to task $\mathcal{T}_i$.  The relatedness among $K$ tasks imply that the structure of $\mathbf{W}$ is low rank or approximately low rank~\cite{DBLP:journals/siamjo/PongTJY10}.
\subsubsection{Subspace Clustering~(SC)}
Given a set of data points approximately drawn from a union of multiple subspaces, the goal of SC is to partition the data points into their respective subspaces. To this end, Liu et al.~\cite{DBLP:journals/pami/LiuLYSYM13} propose a low rank representation model, which seeks a low rank matrix $\mathbf{W}$ that consists of the candidates of all data points in a given dictionary $\mathbf{D}$. A general morel for SC using LRR is:
\begin{equation}\label{eq:sc}
  \min_{\mathbf{W}} \|\mathbf{D}\mathbf{W}-\mathbf{X}\|_{2,1} +\lambda \|\mathbf{W}\|_r\,,
\end{equation}
where the learnt low rank matrix $\mathbf{W}$ can be seen as a rough similarity matrix, and the final partitioning result can be obtained by conducting spectral clustering with a refined similarity matrix $\mathbf{Z}=\frac{|\mathbf{W}|+|\mathbf{W}^T|}{2}$. A significant variant of Eq.~\eqref{eq:sc} is LatLRR~\cite{liu2011latent}, which jointly learns the low rank affine matrix $\mathbf{W}$ and the low rank basis matrix $\mathbf{L}$ using a model of the form:
\begin{equation}
\begin{split}
\min_{\mathbf{W},\mathbf{L}} \|\mathbf{W}\|_r+ \|\mathbf{L}\|_r+ \lambda \|\mathbf{X}-\mathbf{X}\mathbf{W}-\mathbf{L}\mathbf{X}\|_1 \,.
\end{split}
\end{equation}
LatLRR can be considered as a combination of low-rank representation and Inductive Robust Principal Component Analysis~\cite{bao2012inductive}. Such a model and its variants~\cite{zhang2014similarity,zhang2019adaptive,zhang2019robust} have been widely used to deal with computer vision tasks such as image recovery and denoising.

\subsubsection{Summarization}
 For each task mentioned above, there are a great many of algorithms have been developed. For instance, RPCA~\cite{DBLP:conf/kdd/SunXY13,DBLP:journals/pami/OhTBKK16,DBLP:journals/ijcv/GuXMZFZ17}, MC~\cite{DBLP:journals/tit/CandesT10,DBLP:conf/ijcai/NieHH17,DBLP:journals/corr/abs-1710-02004,DBLP:journals/corr/LinCM10,DBLP:conf/cvpr/LuTYL14,DBLP:journals/cacm/CandesR12,chen2016matrix}, MTL~\cite{DBLP:journals/siamjo/PongTJY10,DBLP:conf/aaai/HanZ16,DBLP:journals/pami/ZhenYHL18,DBLP:conf/kdd/ChenZY11,DBLP:conf/icml/JiY09,DBLP:journals/tnn/DingF18,nie2018calibrated,kong2018probabilistic} and SC~\cite{DBLP:conf/ijcai/NieH16,liu2019robust,DBLP:journals/pami/LiuLYSYM13,DBLP:journals/pami/YinGL16,DBLP:journals/tnn/ZhangXSY18,peng2015subspace,zhang2017robust}. Nevertheless, the main differences between these algorithms are loss function or regularization term. A short discussion for loss functions widely used in machine learning can be found in~\cite{nie2018investigation}. The details over the regularization term can be found in Sect.~\ref{sec:2.1}.

Furthermore, it is worth noting that in addition to the models mentioned above, LRR has achieved success on other machine learning tasks. Examples include Component Analysis~\cite{DBLP:conf/cvpr/SagonasPLZ17,DBLP:journals/pami/PanagakisNZP16},  Compressive Sensing~\cite{DBLP:journals/tip/DongSLMH14}, Multi-View Learning~\cite{gao2015multi,DBLP:conf/aaai/LiuLTXW15,DBLP:conf/aaai/XiaPDY14,DBLP:conf/ijcai/WangZWLFP16,ding2018dual,li2017low}, Transfer Learning~\cite{DBLP:conf/aaai/DingSF14,DBLP:journals/ijcv/ShaoKF14}, Spectral Clustering~\cite{chen2014clustering}, Dictionary Learning~\cite{ren2020learning}, Metric Learning~\cite{DBLP:conf/kdd/HuoNH16}, Feature Extraction~\cite{zhang2017robust,wang2019robust} and so on.
Indeed, most of them, including LRR and MTL, are derived from RPCA and MC.
Hence, the recent progress achieved in RPCA and MC may be useful for further improving the performance of existing algorithms.
\begin{table}[]
\small
\centering
\caption{Practical issues solved by using Low Rank Regularization. More details with respect to constructed models and corresponding optimization methods can be found in references.}
\label{my-label}
\begin{tabular}{|l|l|c|}
\hline
                                    Application & Regularizer $\mathcal{R}(\mathbf{X})$ & Optimization \\ \hline\hline
  Face Analysis~\cite{xue2018side,DBLP:conf/iccv/XuePZ17,DBLP:journals/ijcv/SagonasPZP17,DBLP:conf/cvpr/SagonasPLZ17} & $\|\mathbf{X}\|_*$            &     ADMM                \\ \hline
                                    Person Re-Identification~\cite{su2018multi} &   $\|\mathbf{X}\|_*$          &    others                \\ \hline
                                    Visual Tracking~\cite{DBLP:journals/ijcv/SuiTZW18,DBLP:journals/ijcv/SuiZ16} &   $\|\mathbf{X}\|_*$          &   others               \\ \hline
                                    $3$D Reconstruction~\cite{agudo2018image,DBLP:journals/ijcv/DaiLH14,DBLP:conf/cvpr/ZhuHTL14} &   $\|\mathbf{X}\|_*$    &    ADMM               \\ \hline
                                    Image denoising~\cite{DBLP:journals/ijcv/GuXMZFZ17,DBLP:conf/iccv/ChangYZ17,DBLP:journals/tip/HuangDXSB17,DBLP:conf/iccv/WangT13}       & $\sum_{i=1}^{k} w_i\sigma_i$          &     ADMM \\ \hline
                                    Structure Recovery~\cite{DBLP:conf/iccvw/GS17}       &   $\sum_{i=1}^{r} w_i\sigma_i$          &    others \\\hline
                                    Video Desnowing and Deraining~\cite{DBLP:conf/cvpr/RenTHCT17} & $\|\mathbf{X}\|_*$            &    AM                \\ \hline
                                    Salient Object Detection~\cite{DBLP:journals/pami/PengLLHXM17,DBLP:conf/cvpr/ShenW12,DBLP:conf/bmvc/ZouKLR13,gao2014block} & $\|\mathbf{X}\|_*$            &     ADMM               \\ \hline
                                     Face Recognition~\cite{DBLP:journals/pami/YangLQTZX17,DBLP:conf/cvpr/LezamaQS17,li2014learning}       &   $\|\mathbf{X}\|_*$         &     ADMM \\\hline

                                    High Dynamic Range Imaging~\cite{DBLP:journals/pami/OhLTK15,DBLP:journals/pami/OhTBKK16}       &   $\|\mathbf{X}\|_*$, $\sum_{i=k+1}^{r}\sigma_i$         &     ADMM \\ \hline
                                    Head Pose Estimation~\cite{DBLP:conf/aaai/ZhaoDF16}       &   $\|\mathbf{X}\|_*$         &     ADMM \\\hline
                                    Moving Object Detection~\cite{DBLP:conf/iccv/ShakeriZ17}       &   $\|\mathbf{X}\|_*$         &     ADMM \\\hline
                                    Reflection Removal~\cite{DBLP:conf/cvpr/HanS17}       &   $\|\mathbf{X}\|_*$         &     others \\\hline
                                    Zero-Shot Learning~\cite{DBLP:conf/cvpr/DingSF17}       &   $\sum_{i=r+1}^{k}\sigma_i$         &     others \\\hline
                                    Speckle removal~\cite{DBLP:conf/cvpr/ZhuFBH17}       &   $\sum_{i=r+1}^{k}w_i\sigma_i$         &     ADMM \\\hline
                                    Image Completion~\cite{DBLP:conf/eccv/LiLXSG16}       &   $\sum_{i=r+1}^{k}w_i\sigma_i$         &     ADMM \\\hline
                                    Image Matching~\cite{DBLP:conf/iccv/ZhouZD15}       &   $\frac{1}{2}(\|\mathbf{U}\|_F^2+\|\mathbf{V}\|_F^2)$        &     ADMM \\\hline
                                    Video Segmentation~\cite{li2015sold}       &   $\|\mathbf{UV}\|_F^2$        &     others \\\hline
                                    Image alignment~\cite{DBLP:journals/ijcv/PengZYM18,DBLP:conf/cvpr/SagonasPZP14,DBLP:journals/pami/PengGWXM12,DBLP:conf/cvpr/ZhaoCW11}&   $\|\mathbf{X}|_*$        &     ADMM \\\hline
                                    Image Restoration~\cite{DBLP:conf/iccv/LiCZGT15}&   $\|\mathbf{X}\|_*$        &     ADMM \\\hline
                                    Image Classification~\cite{DBLP:journals/pami/CabralTCB15,DBLP:conf/iccv/ZhangGLXA13,DBLP:journals/tip/ZhangLZZY16,DBLP:conf/aaai/ZhuJWWCLH17,DBLP:conf/aaai/JiangGP14,DBLP:conf/cvpr/ZhangJD13}&   $\|\mathbf{X}\|_*$          &  ADMM   \\ \hline
                                    AAM fitting~\cite{DBLP:conf/iccv/ChengSSL13}&   $\|\mathbf{X}\|_*$          &  ADMM\\ \hline
                                    Image Segmentation~\cite{cheng2011multi}&   $\|\mathbf{X}\|_*$          &  ADMM \\ \hline
                                    Motion Segmentation~\cite{DBLP:conf/iccv/LiuY11}&   $\|\mathbf{X}\|_*$          &  ADMM\\ \hline
                                    Colorization~\cite{DBLP:conf/aaai/YaoJ15,DBLP:conf/aaai/WangZ12}&   $\|\mathbf{X}\|_*$          &  ADMM\\ \hline
                                    photometric stereo~\cite{DBLP:conf/hsi/FanLQWDY16} & $\|\mathbf{X}\|_*$ & ADMM   \\ \hline
                                    Textures~\cite{DBLP:journals/ijcv/ZhangGLM12}&   $\|\mathbf{X}\|_*$          &  ADMM\\ \hline
                                    Behavior Analysis~\cite{DBLP:journals/ijcv/GeorgakisPP18}    & $\|\mathbf{X}\|_{Sp}$            &     ADMM                 \\ \hline
                                    Heart Rate Estimation~\cite{DBLP:conf/cvpr/TulyakovARYCS16}       &    $\|\mathbf{X}\|_*$         &     ADMM                \\ \hline
                                    Text Models~\cite{DBLP:conf/nips/Shi13}       &    $\|\mathbf{X}\|_*$         &     others                \\ \hline
                                    Rank Aggregation~\cite{DBLP:conf/aaai/PanLLTY13}       &    $\|\mathbf{X}\|_*$         &     ADMM                \\ \hline
                                    Deep Learning~\cite{DBLP:conf/eccv/DingSF16,lezama2018ole,ding2019deep} & $\|\mathbf{X}\|_*$ & SGD \\ \hline

\end{tabular}
\setlength{\abovecaptionskip}{-10.cm}
\setlength{\belowcaptionskip}{-5.cm}
\end{table}
\subsection{Applications of LRR in practical tasks}\label{sec:2.2}
A basic assumption in machine learn models mentioned above is that the columns
or rows of target matrix are relevant. Such an assumption
is often satisfied in real data. We provide a summarization for practical issues solved by using LRR models in Table~\ref{my-label}. Note that the models used in solving these issues are partly derived from the machine learning models mentioned above. Next, we describe the details of some representatives among these tasks.
\subsubsection{Video background subtraction}
Video background subtraction~\cite{candes2011robust} aims to recover the background component of a video data. Given a video with $n$ gray images of size $wh$, we vectorize each image as a vector $\mathbf{x}\in\mathbb{R}^{wh}$. Stacking all vectors, we can obtain a matrix $\mathbf{X}\in \mathbb{R}^{wh\times n}$. The problem of video background subtraction can be solved by solving a RPCA model of the form:
\begin{equation}\label{eq:bc}
\min_{\mathbf{L},\mathbf{E}} \ \ \|\mathbf{E}\|_1 + \lambda\|\mathbf{L}\|_r \,,\ \
      s.t. \ \ \mathbf{X} = \mathbf{L}+ \mathbf{E}\,,
\end{equation}
where $\mathbf{L}$ is a low rank matrix, recording the background components of video, while $\mathbf{E}$ is a sparse matrix, recording the foreground components of video. In order to achieve better splitting results, numerous improvements for Eq.~\eqref{eq:bc} have been made. In~\cite{DBLP:journals/ijcv/GuXMZFZ17}, the nuclear norm is replaced by non-convex regularizers to achieve better low rank approximation. In order to cope with the nonrigid motion and dynamic background, a DECOLOR method~(DEtecting Contiguous Outliers in the LOw-rank Representation) is proposed in~\cite{DBLP:journals/pami/ZhouYY13}. Recently, Shakeri et al.~\cite{DBLP:conf/iccv/ShakeriZ17} construct a low-rank and invariant sparse decomposition model to reduce the effect caused by various illumination changes. For background subtraction is one of the indices to evaluate the performance of RPCA algorithms, it has received a great many of attentions, and a comprehensive survey can be found in~\cite{DBLP:journals/csr/BouwmansSJJZ17}~\footnote{BS library: https://github.com/andrewssobral/lrslibrary}.
\subsubsection{Recommender System}
The goal of recommender system is leveraging some prior information on user behavior to predict user preferences~\cite{koren2009matrix}. The Collaborative
Filtering~(CF), as exemplified by low rank matrix learning methods, have proven successful in dealing with this task. In order to predicate the preferences of users, one would like to recover the missing values of a rating matrix $\mathbf{M}$. Such a task can be solved by using a matrix completion model of the form:
 \begin{equation}\label{eq:rmc-2}
        \min_{\mathbf{X}}\ \ \|\mathcal{P}_\Omega(\mathbf{X}-\mathbf{M})\|_F^2 + \lambda \|\mathbf{X}\|_r \,, \\
\end{equation}
where $\mathbf{X}$ denotes the complete rating matrix. This task is a significant metric for evaluating the performance of matrix completion models and has attached much attention in recent years. The main lines of research are rank minimization~\cite{yao2016efficient,DBLP:journals/corr/abs-1708-00146,li2019both} and matrix factorization~\cite{xu2020multi,li2019zero}. Particularly, the latter often assumes that the rank $r$ of target matrix is known and replace $\mathbf{X}\in\mathbb{R}^{n\times m}$ with two new variables $\mathbf{U}\in\mathbb{R}^{m\times r}$ and $\mathbf{V}\in\mathbb{R}^{r\times n}$, where $\mathbf{X}=\mathbf{UV}$. An overview for solving  such a classes of non-convex optimization problems has been provide in~\cite{chi2019nonconvex}.
\subsubsection{Image denoising}
Image denoising, in essence, involves estimating the latent clean image from an noisy image, which is a fundamental problem in low level vision. The success of several state-of-the-art image denoising algorithms~\cite{DBLP:journals/tip/DabovFKE07,DBLP:conf/cvpr/GuZZF14} is based on the exploitation of image nonlocal self-similarity, which refers to the assumption that for each local patch in a natural image, one can find some similar patches to it. Given an noise image, Gu et al.~\cite{DBLP:conf/cvpr/GuZZF14,DBLP:conf/accv/WangZL12} propose to vectorize all similar patches as column vectors and stacking them as a noisy matrix $\mathbf{M}$. Further, the clean patches can be recovered by using a RPCA model of the form:
\begin{equation}\label{eq:id}
\begin{split}
    \min_{\mathbf{X}} &\  \   \|\mathbf{X}-\mathbf{M}\|_F^2 + \lambda\|\mathbf{X}\|_r\,,\\
\end{split}
\end{equation}
where  $\mathbf{X}$ denotes the clean patches we aim to recover. Similar models with different regularizers have been proposed in~\cite{DBLP:journals/tip/HuangDXSB17}, and a different model with similar regularizer has been developed in~\cite{yair2018multi,DBLP:conf/iccv/XuZ0F17}.

\subsubsection{Image alignment}
Image Alignment~(IA) involves  transforming various images into a common coordinate system. Given a set of images, Peng et al.~\cite{DBLP:journals/pami/PengGWXM12}  vectorize each image as a vector $\mathbf{m}$ and stack all vectors as a matrix $\mathbf{M}$. The alignment images $\mathbf{X}$ can be recovered by using a model of the form:
\begin{equation}\label{eq:IA}
\textstyle
    \min_{\mathbf{X},\mathbf{E},\tau} \  \   \|\mathbf{E}\|_1 + \lambda\|\mathbf{X}\|_r \ \ s.t. \ \ \mathbf{M} \circ \tau= \mathbf{X}+\mathbf{E}\,,
\end{equation}
where $\tau$ denotes a transformation, $\mathbf{E}$ denotes the noise or the differences among images. There are two variants for this method~\cite{DBLP:conf/iccv/ChengSSL13,DBLP:journals/ijcv/PengZYM18}, but both of them select the nuclear norm to serve as regularizer and the ADMM method to optimize the corresponding models.

\subsubsection{Non-rigid structure from motion}
Non-rigid Structure From Motion~(NSFM) is a powerful tool for recovering the $3$D structure and pose of an object from a single image, which is well known to be an ill-posed problem in literature due to the non-rigidity. Assuming that the  nonrigid $3$D shapes lie in a single low dimensional subspace, Dai et al.~\cite{DBLP:journals/ijcv/DaiLH14} propose to incorporate low rank prior into NSFM. Specifically, the model for inferring the $3$D coordinates is:
\begin{equation}\label{eq:NSFM}
 \min_{\mathbf{X},\mathbf{E}}  \  \ \|\mathbf{X}\|_r+\lambda\|\mathbf{E}\|_1 \ \ s.t. \  \ \mathbf{W} = \mathbf{R}\mathbf{X}^\#+\mathbf{E}\,,
\end{equation}
where $\mathbf{W}$ consists of the $2$D projected coordinates of all data points, $\mathbf{X}$ denotes the $3$D coordinates we aim to recover. In addition, the definitions of $\mathbf{R}$ and $\mathbf{X}^\#$ can be found in~\cite{DBLP:journals/ijcv/DaiLH14}.  Obviously, such a formulation is derived from RPCA. Subsequently, in order to cope with the complex nonrigid motion that $3$D shapes lie in a union of multiple subspaces rather than a single subspace, Zhu et al.~\cite{DBLP:conf/cvpr/ZhuHTL14} propose a subspace clustering based model. This model learns a $3$D structure matrix $\mathbf{X}$ and an affinity matrix $\mathbf{Z}$, simultaneously. And $\mathbf{Z}$ is used to obtain the final clustering result. Besides, a more complicated case that the $2$D point tracks contain multiple deforming objects is studied in ~\cite{DBLP:conf/cvpr/AgudoM17,DBLP:journals/corr/KumarDL17,agudo2018image}.
\subsubsection{Deep learning with LRR}
Deep learning is a powerful tool to tackle various tasks, which leverage  Deep Neural
Networks~(DNN) to integrate features learning and weights learning into a unified framework. Recently, numerous studies~\cite{DBLP:conf/eccv/DingSF16,lezama2018ole,xu2018trained,piao2019double} propose to introduce the low rank regularization into deep learning. Similarly, all these methods can be grouped into two categories including low rank features learning~\cite{zhong2019unsupervised,lezama2018ole} and low rank weights learning~\cite{piao2019double}. To learn large-margin deep features for classification problem, Lezama et al.~\cite{lezama2018ole} propose an Orthogonal Low-rank Embedding~(OLE) method, which enforces the intra-class embeddings falling in a subspace aligned with the corresponding weight vector, while embeddings of different classes be orthogonal to each other. The OLE loss is:
\begin{equation}\label{ole}
  \min_{\mathbf{X}} = \sum_{c=1}^{C} \max(\Delta,\|\mathbf{X}_c\|_r) - \|\mathbf{X}\|_r\,,
\end{equation}
where $\mathbf{X}$ denotes the deep embeddings of data, while $\mathbf{X}_c$ is a sub-matrix of $\mathbf{X}$, which consists of the embeddings from the class $c$. Besides, $\Delta$ is a parameter to avoid the collapse of embeddings to zero. By combining OLE loss and stand cross-entropy loss, the author achieve a better result in image classification task. Subsequently, Zhu et al.~\cite{zhu2019stop} generalize OLE loss to serve as a regularization term to improve the generalization ability of DNN.

Low rank weights learning refers to train a DNN whose weights are low rank or approximately low rank, which plays on significant role in the task of compressing DNN model~\cite{zhang2015efficient}.
In order to compress the size of trained network, Xu et al.~\cite{xu2018trained} integrates low rank approximation~(over weights) into the procedure of training deep neural networks. Such a strategy can guarantee that the trained network has a low-rank structure in nature, and eliminate the error caused by fine-tuning. Besides, Piao et al.~\cite{piao2019double} leverage LRR to deal with the deep subspace clustering problem, where learnt low rank weight matrix is consistent with the affine matrix of traditional subspace clustering. Recently, Sanyal et al.~\cite{sanyal2019stable} propose to use Stable Rank Rormalization~(SRN) to improve the generalization performance of DNN.
\subsection{Summarization}
In addition to the tasks mentioned above, LRR have been applied into other tasks, including Visual Tracking~\cite{DBLP:journals/ijcv/SuiTZW18,DBLP:journals/ijcv/SuiZ16,DBLP:journals/ijcv/Zhang0AYG15}, salient object detection~\cite{DBLP:journals/pami/PengLLHXM17,DBLP:conf/cvpr/ShenW12,DBLP:conf/bmvc/ZouKLR13,gao2014block}, face analysis~\cite{xue2018side,DBLP:journals/ijcv/SagonasPZP17,DBLP:conf/cvpr/SagonasPLZ17}
and so on.  A comprehensive summarization for these tasks are reported in Table~\ref{my-label}. Indeed, most of these algorithms are derived from the machine learning models mentioned above. But, observing Table~\ref{my-label}, one can find that in these tasks nuclear norm often serves as the relaxation function, while ADMM generally serves as the optimization method. And researchers do not pay enough attention to the recent progress made in non-convex relaxation and optimization. Such an issue is also the motivation of this review. We prefer to  promote the application of non-convex
relaxations in practical issues. Next, we experimentally verify the advantage of non-convex relaxations over nuclear norm.

\section{Comparison between different relaxations to LRR} \label{sec:5}
For the main goal of this paper is promoting the application of non-convex relaxations in solving practical issues, in this section we would like to take a comprehensive investigation for several non-convex relaxations.

A huge number of LRR models have been developed for addressing various data analysis tasks in the past decade. However, the loss function and regularization term are the main differences between these models. Particularly, the selection of loss function depends on the problem we aim to solve. Taking a comprehensive comparison for all tasks beyond the scope of our ability.  In this work, we conduct our investigation on image denoising task. As presented in~\cite{DBLP:journals/ijcv/GuXMZFZ17}, the core in this task is solving a LRR model as follows:
\begin{equation}\label{eq:image denoising}
  \min_{\mathbf{X}}  \|\mathbf{X}-\mathbf{M}\|_F^2+\lambda\sum_{i=1}^{k}f(\sigma_i(\mathbf{X}))\,.
\end{equation}
The reasons of using such a task include:
\begin{enumerate}
  \item As discussed above, the problem~\eqref{eq:image denoising} is a basic problem that we cannot avoid when dealing with complicated LRR models;
  \item The model~\eqref{eq:image denoising} has only one parameter needed to be tuned. Such that the results can be analysed conveniently.
\end{enumerate}
    The relaxations to rank-norm  include: Weighted Nuclear Norm~(WNN), Log Nuclear Norm~(LNN), Truncated Nuclear Norm~(TNN), and Schatten-p norm with $p=1$, $p=0.5$, $=0.01$, respectively. Particularly, Nuclear Norm~(NN) is equivalent to the Schatten-p norm with $p=1$. For each regularizer we can obtain the solution of problem~\eqref{eq:image denoising} directly via SVT or GSVT. Here, we abandon the Capped Nuclear Norm, ETP, Laplace, Geman and so on, for all of them have additional parameters needed to be tuned.

\begin{figure}[t]
	\centering
	\subfloat{\includegraphics[width=0.24\linewidth]{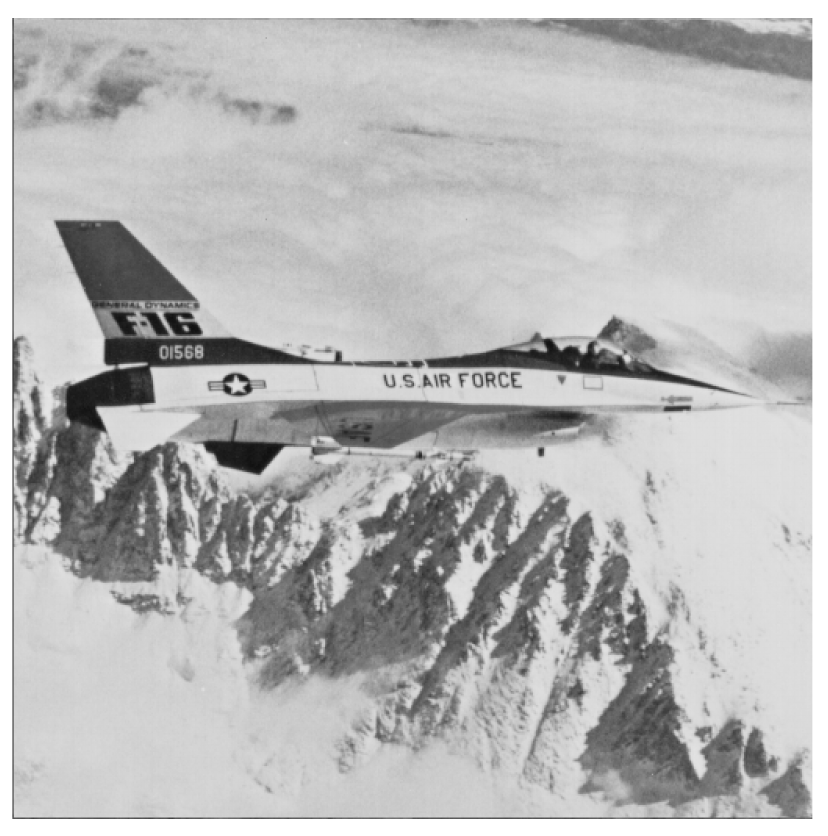}}
     \subfloat{\includegraphics[width=0.24\linewidth]{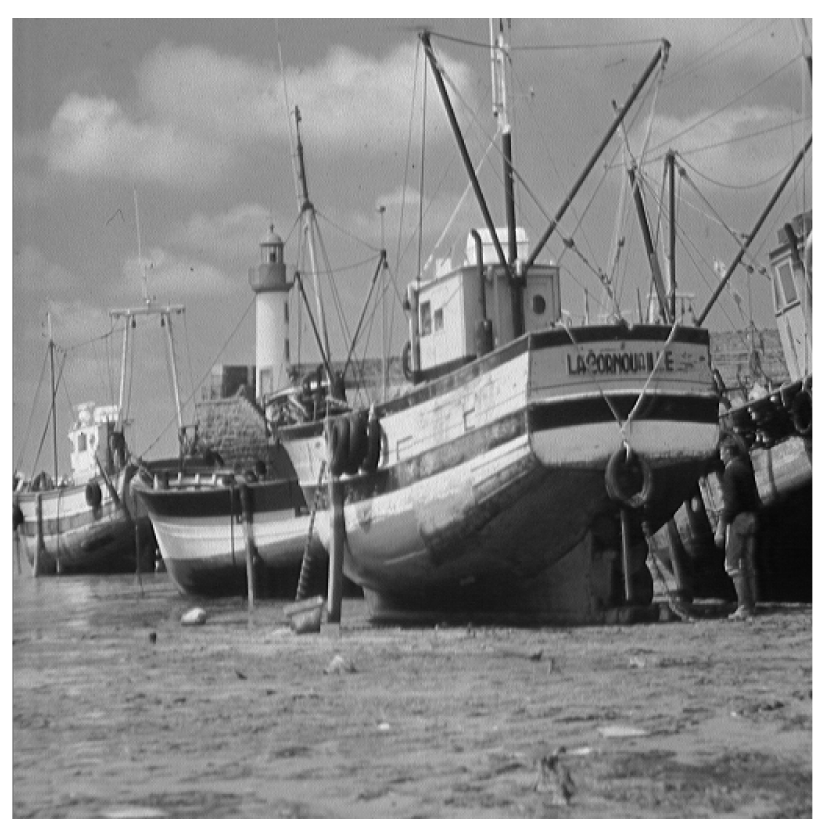}}
     \subfloat{\includegraphics[width=0.24\linewidth]{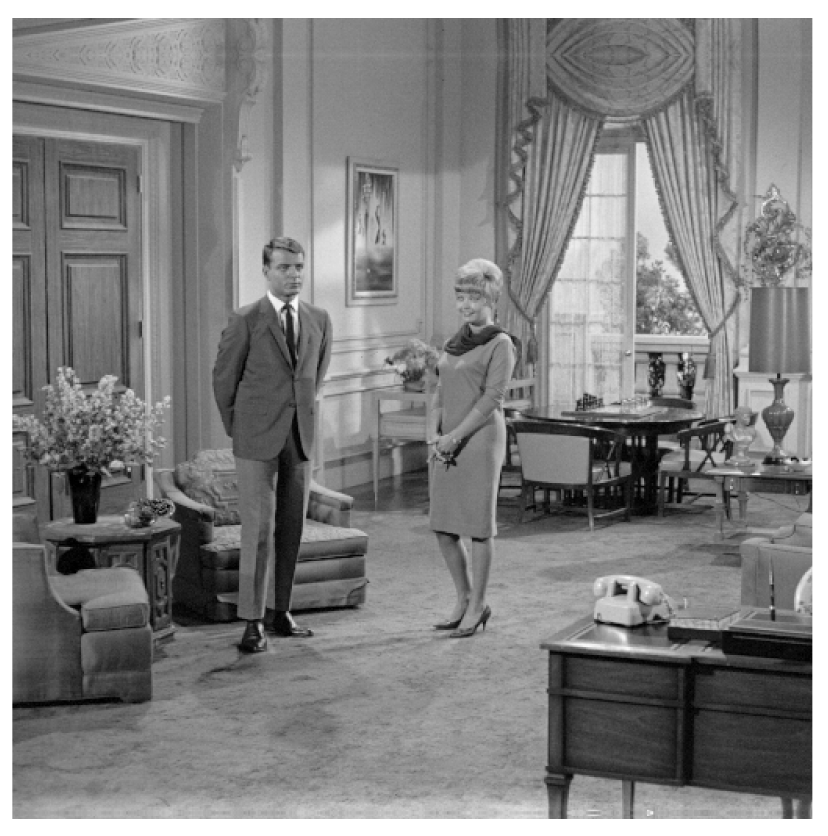}}
     \subfloat{\includegraphics[width=0.24\linewidth]{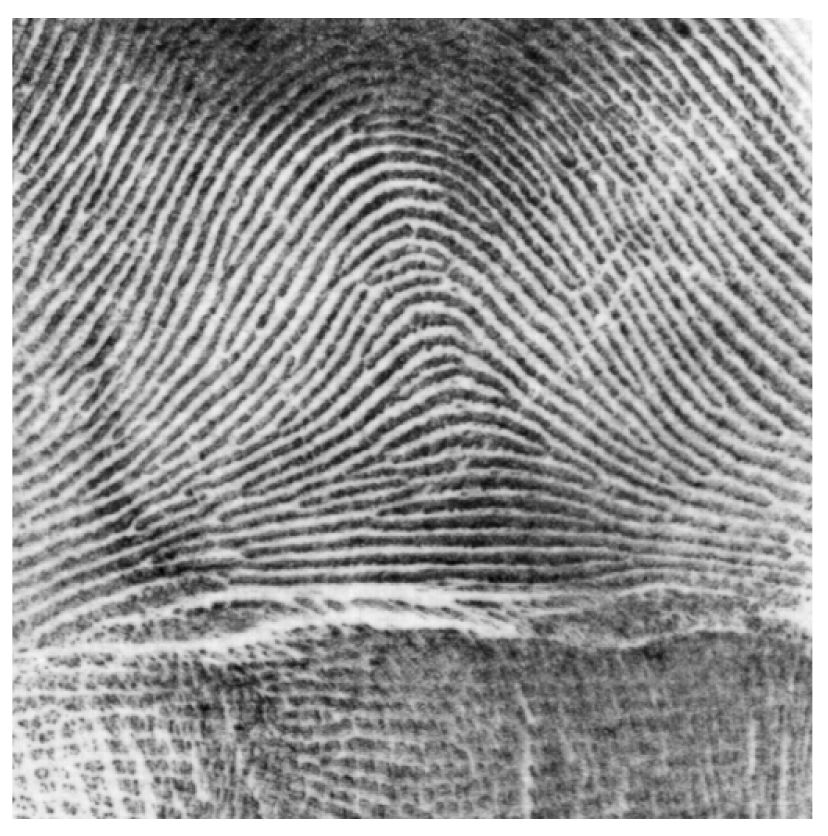}}
     \hfil
     \subfloat{\includegraphics[width=0.24\linewidth]{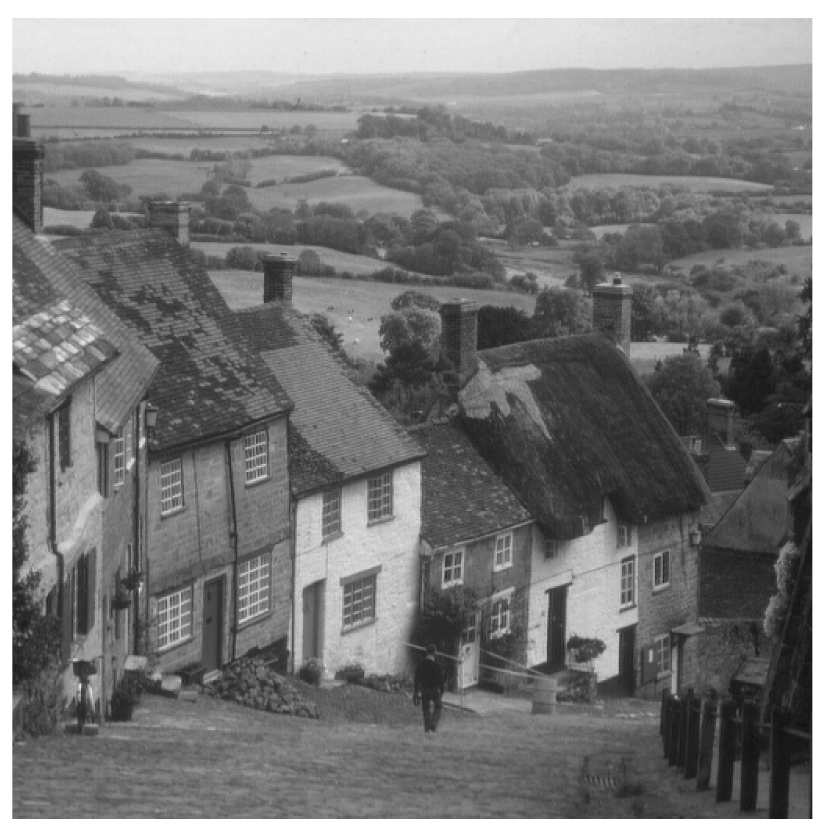}}
     \subfloat{\includegraphics[width=0.24\linewidth]{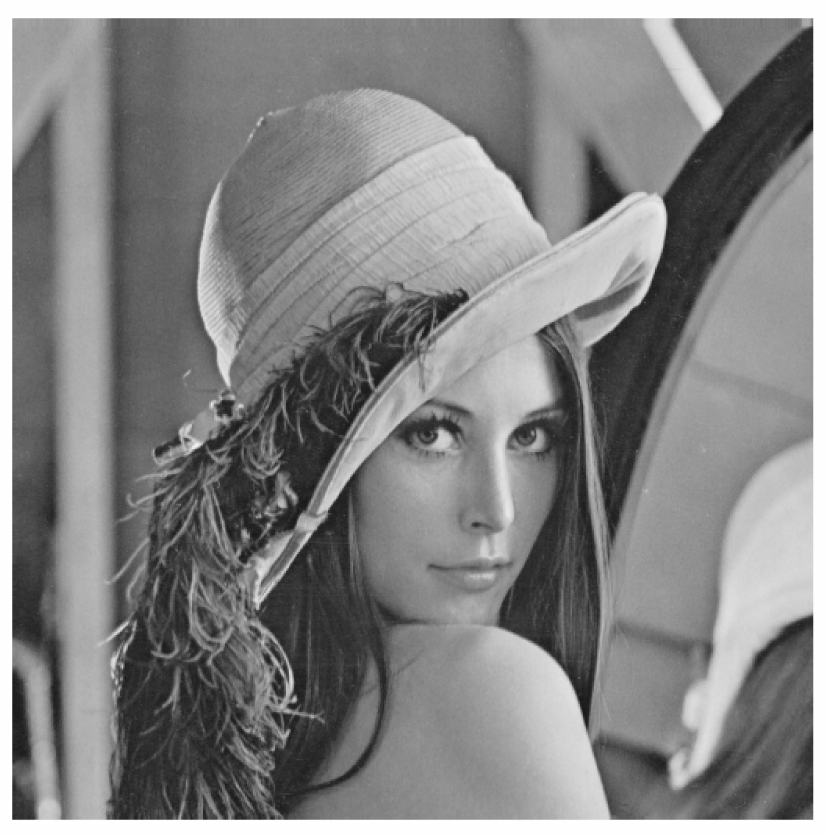}}
     \subfloat{\includegraphics[width=0.24\linewidth]{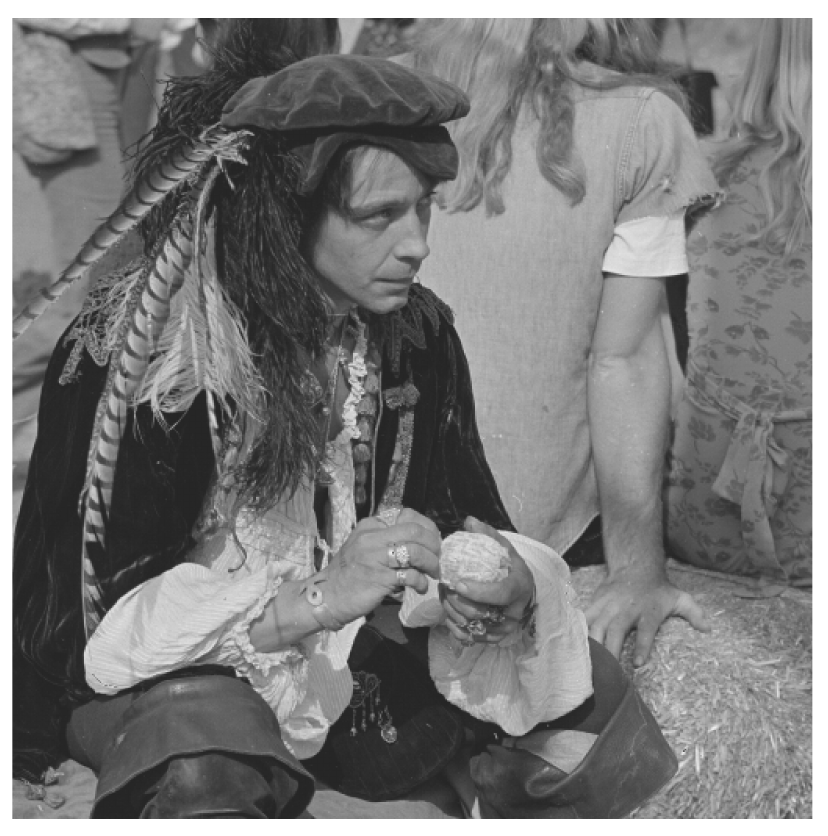}}
     \subfloat{\includegraphics[width=0.24\linewidth]{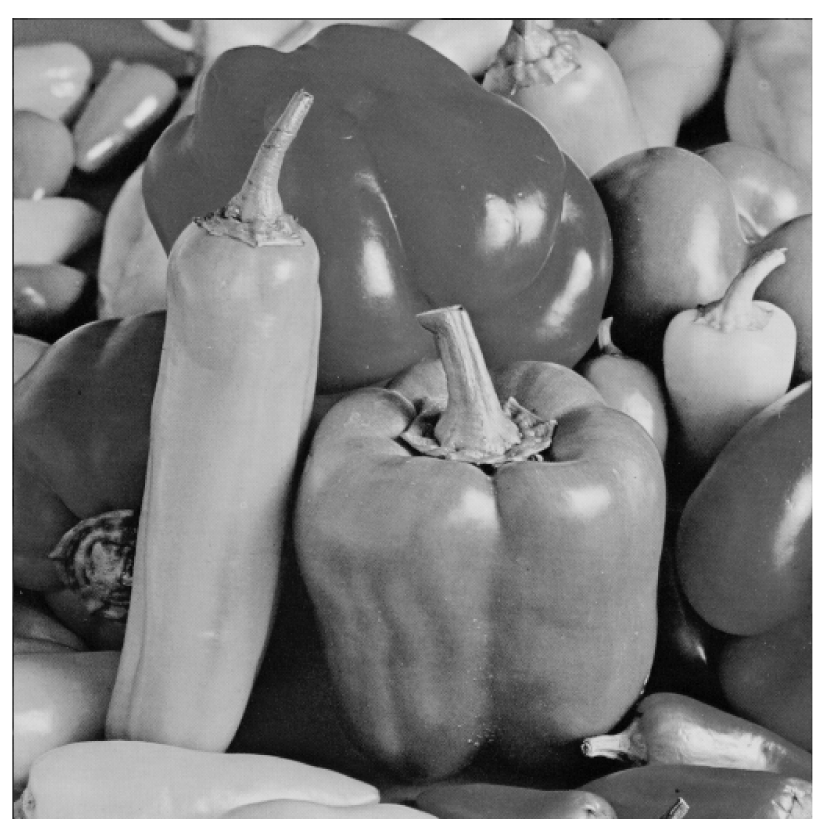}}
	\caption{The $8$ test images used in this paper.}
	\label{Fig:original}
\end{figure}
\begin{figure*}[t]
	\centering
	\subfloat[$\tau=0.01$]{\includegraphics[width=0.33\linewidth]{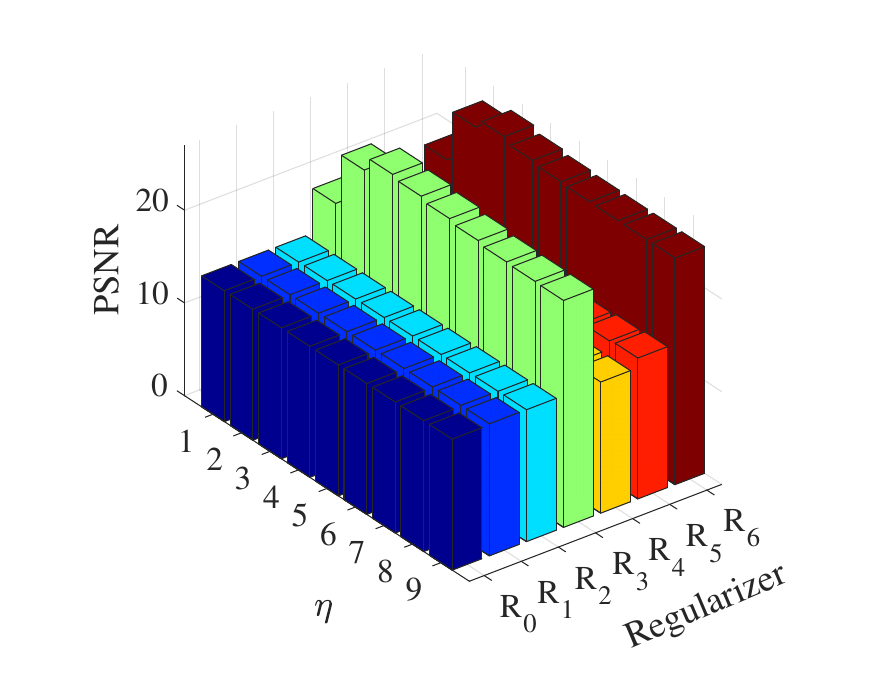}
		\label{Fig1a}}
   \subfloat[$\tau=0.1$]{\includegraphics[width=0.33\linewidth]{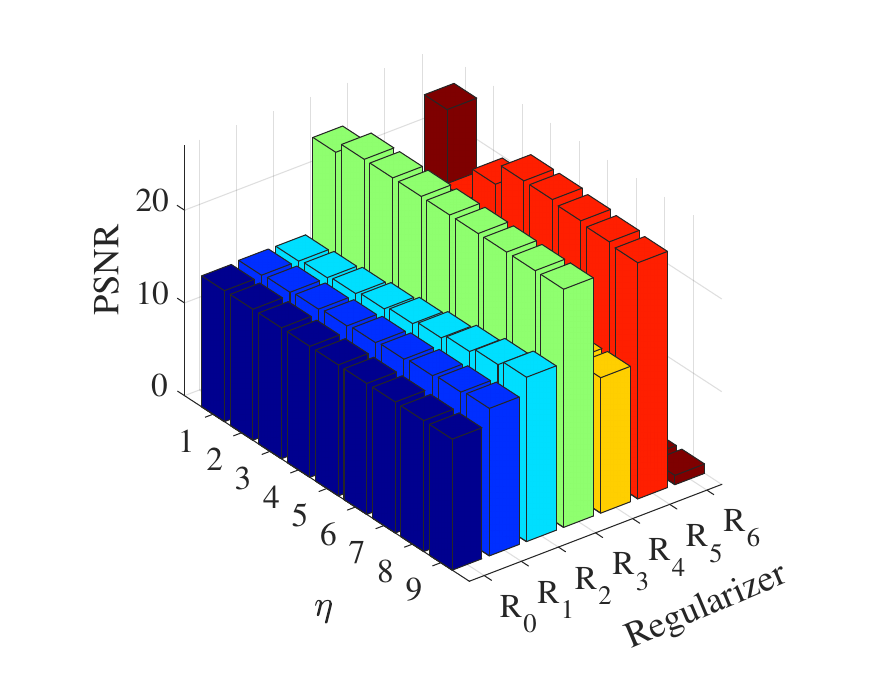}
		\label{Fig1b}}
	\subfloat[$\tau=1$]{\includegraphics[width=0.33\linewidth]{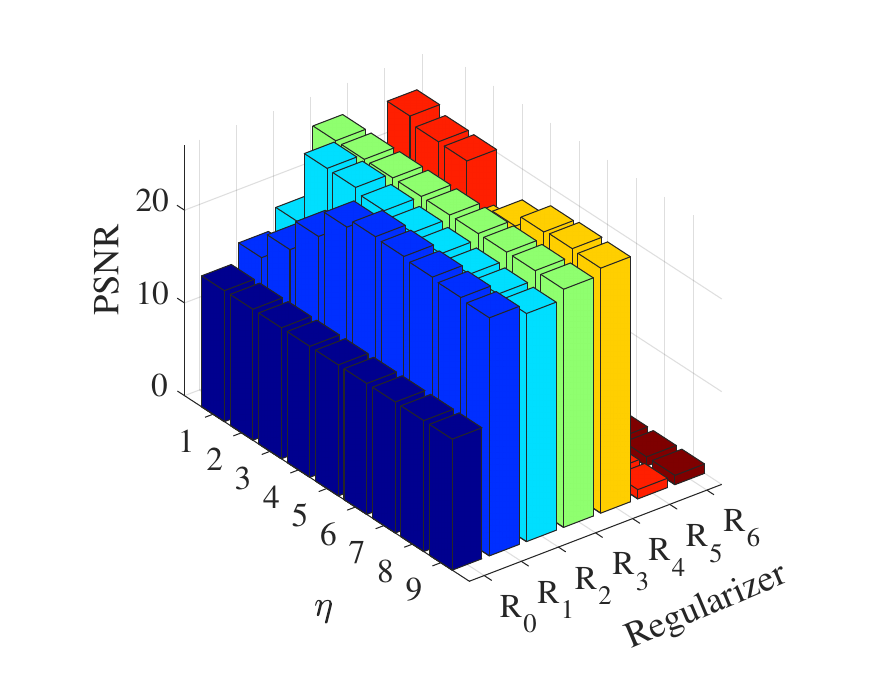}
		\label{Fia1c}}
	\caption{A comprehensive comparison between different regularizers on image \emph{Peppers}. $R_0$ refers to the PSNR of input noisy image. $\{R_1,R_2,\ldots,R_6\}$ refer to the regularizers WNN, LNN, TNN, $S_p$ with $p=0.1$, $S_p$ with $p=0.5$, and NN~(Nuclear Norm, $S_p$ with $p=1$), respectively. Note that all relaxations, except for TNN, will shrink all singular values to $0$ when $\lambda$ takes a large value.}
	\label{Fig:2}
\end{figure*}
\begin{figure*}[t]
	\centering
	\subfloat[$\tau=0.01$]{\includegraphics[width=0.33\linewidth]{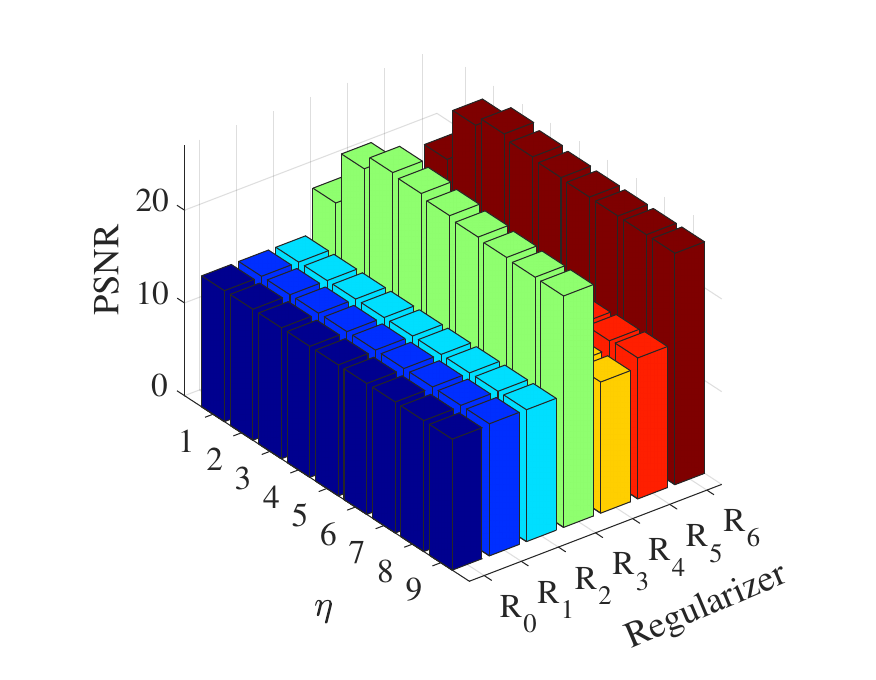}
		\label{Fig1a}}
   \subfloat[$\tau=0.1$]{\includegraphics[width=0.33\linewidth]{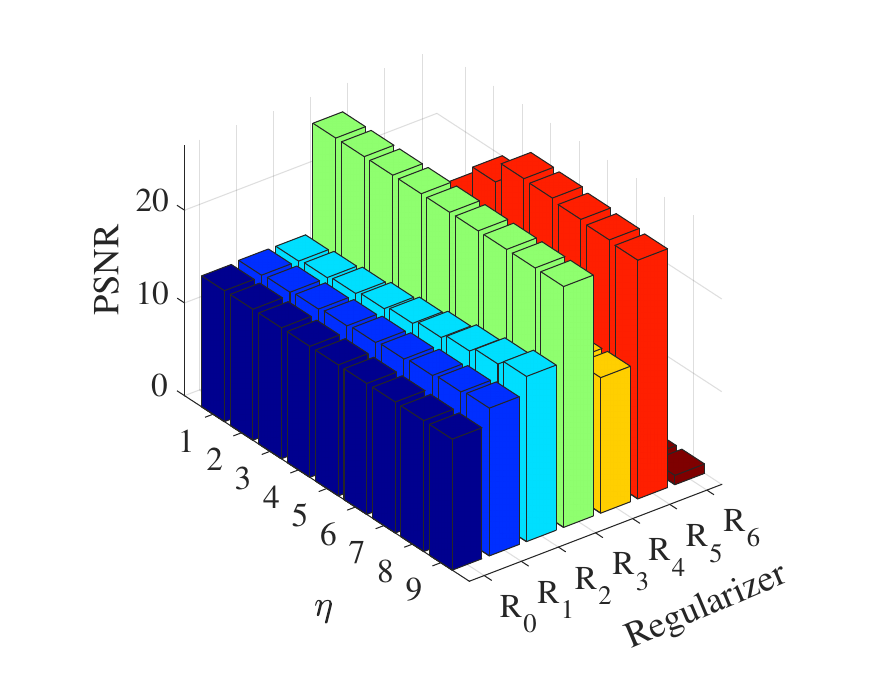}
		\label{Fig1b}}
	\subfloat[$\tau=1$]{\includegraphics[width=0.33\linewidth]{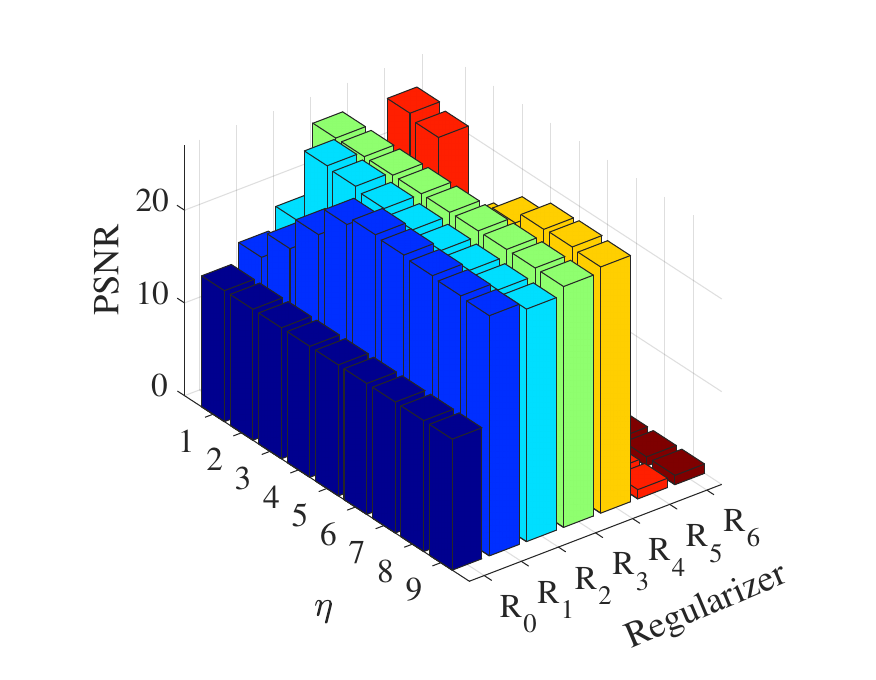}
		\label{Fia1c}}
	\caption{A comprehensive comparison between different regularizers on image \emph{Lena}. $R_0$ refers to the PSNR of input noisy image.}
	\label{Fig:3}
\end{figure*}

\subsection{Experimental setting}
We select $8$ widely used images with size $256\times 256$ to evaluate the competing regularizers. The thumbnails are shown in Fig~\ref{Fig:original}. Similarly, we corrupt the original image by Gaussian noise with distribution $\mathcal{N}(0,\sigma^2)$.

We use the codes provided by Gu et al.~\cite{DBLP:journals/ijcv/GuXMZFZ17}~\footnote{https://sites.google.com/site/shuhanggu/home}. For each patch, in~\cite{DBLP:journals/ijcv/GuXMZFZ17} the authors run $K$ iterations of this approximation process to enhance the quality of denoising. In this paper we fix $K=1$ and other parameters as authors suggested to avoid introducing additional parameters. More implementation details over experiments can be found in~\cite{DBLP:journals/ijcv/GuXMZFZ17}.

For all regularizers except for WNN we set $\lambda = \tau \eta \sqrt{n_p}\sigma^2$, where $n_p$ is the number of similar parts, $\sigma$ represents the noise level, $\tau$ controls the large scale range of varying $\lambda$, and $\eta$ controls the small scale range of varying $\lambda$. For WNN, we set $\lambda_w = C*\lambda$, where $C=\sqrt{2}$ as authors suggested. For TNN, we fix $r=5$ for all test images.

\textbf{Comparison~$1$.} In this test, we fix $\sigma=50$ and vary $\tau$ in the set $\mathcal{S}_\tau=\{0.1, 1 ,10\}$, $\eta$ in the set $\mathcal{S}_{\eta}=\{1,2,\ldots,9\}$. We select only \emph{Lena} and \emph{Peppers} two images to test. The PSNR results under different parameters for all competing relaxations are shown in Figure~\ref{Fig:2} and Figure~\ref{Fig:3}, where $R_0$ refers to the PSNR of input noisy image, $\{R_1,R_2,\ldots,R_6\}$ refer to the relaxations WNN, LNN, TNN, $S_p$ with $p=0.1$, $S_p$ with $p=0.5$, and nuclear norm, respectively. According to the definition of SVT and GSVT, we can find that all relaxations will shrink all singular values to zero when $\lambda$ takes a large value~(Under this case, we set $PSNR=1$ for visualization).
Observing the Figure~\ref{Fig:2} and Figure~\ref{Fig:3}, we can achieve the follow conclusions.
\begin{itemize}
  \item Non-convex relaxations consistently outperforms nuclear norm~(convex relaxation) in terms of the best result;
  \item We should select a smaller regularization parameter $\lambda$ for nuclear norm, for it prefers to shrink all singular values to zero when $\lambda$ takes a large value~(see the cases that $\tau>1$ and $\eta>5$). Nevertheless, the shrink for singular values may be insufficient when $\lambda$ takes small value~(see the case that $\tau=0.01$ and $1\leq\eta \leq4$). Such a contradiction limits the performance of nuclear norm;
  \item TNN performs well even when $\lambda$ takes a large value, for it prevents the $r$ largest singular values from being shrinked. Nevertheless, two limitations of it cannot be ignored. First, the value of $r$ must be estimated. Second, the $r$ largest singular values generally carry noise information we want to remove, and preserving them may degenerate the performance of algorithm;
  \item We should select a larger regularization parameter $\lambda$ for non-convex relaxations. When $\lambda$ takes a small value, the shrink on all singular values is vary slight. Hence, the PSNRs of them are very close to original images. Non-convex relaxations can reduce the shrink on larger singular values and enhance the shrink on smaller singular values simultaneously compared  with nuclear norm.
\end{itemize}
In this test, we ignore the numerical differences~(PSNR values) between different non-convex relaxations, because such a difference can be compressed by carefully selecting regularization parameter  for each regularizer.

\begin{table*}[]
\scriptsize
\centering
\caption{Experimental results~(PSNR,dB) on the $8$ test images. Here, WNN-I refers to the algorithm proposed in~\cite{DBLP:journals/ijcv/GuXMZFZ17}, which iteratively conduct the reconstruction process on all patches. The best results~(without considering WNN-I) are highlighted in bold.}\label{tab:res}
\begin{tabular}{|c|c|c|c|c|c|c|c|c|}
\hline
Noise              & Images       & WNN-I  & WNN    & LNN      & TNN ($r=1$)      & $S_p$($p=0.1)$      & $S_p$($p=0.5)$      & Nuclear Norm      \\ \hline
\multirow{8}{*}{$\sigma=10$} & Airplane   & 34.732 & 34.352 & 34.130 & 30.146 & \textbf{34.495} & 31.527 & 30.194 \\ \cline{2-9}
                   & Boat        & 33.727 & 33.186 & 33.309 & 30.023 & \textbf{33.512} & 31.090 & 30.051 \\ \cline{2-9}
                   & Couple      & 33.735 & 33.247 & 33.240 & 30.019 & \textbf{33.551} & 31.006 & 30.036 \\ \cline{2-9}
                   & Fingerprint & 31.069 & 30.907 & 30.696 & 29.420 & \textbf{30.999} & 29.783 & 29.422 \\ \cline{2-9}
                   & Hill        & 33.715 & 33.300 & 33.386 & 30.075 & \textbf{33.629} & 31.101 & 30.101 \\ \cline{2-9}
                   & Lena        & 35.721 & 35.308 & 35.026 & 30.282 & \textbf{35.496} & 31.724 & 30.318 \\ \cline{2-9}
                   & Man         & 33.436 & 32.998 & 33.069 & 29.998 & \textbf{33.329} & 30.963 & 30.022 \\ \cline{2-9}
                   & Peppers     & 35.965 & 35.593 & 35.149 & 30.292 & \textbf{35.736} & 31.714 & 30.323 \\ \hline\hline
\multirow{8}{*}{$\sigma=30$} & Airplane   & 28.749 & 28.077 & 28.200 & \textbf{26.357} & 28.242 & 27.176 & 26.470 \\ \cline{2-9}
                   & Boat        & 27.875 & 27.144 & \textbf{27.413} & 26.013 & 27.273 & 26.588 & 26.045 \\ \cline{2-9}
                   & Couple      & 27.698 & 26.893 & \textbf{27.198} & 25.885 & 27.046 & 26.359 & 25.883 \\ \cline{2-9}
                   & Fingerprint & 24.919 & 24.637 & 24.776 & 24.193 & \textbf{24.790} & 24.221 & 24.117 \\ \cline{2-9}
                   & Hill        & 28.472 & 27.759 & \textbf{28.013} & 26.436 & 27.868 & 27.025 & 26.492 \\ \cline{2-9}
                   & Lena        & 29.929 & 29.060 & 29.167 & 26.970 & \textbf{29.186} & 27.803 & 27.094 \\ \cline{2-9}
                   & Man         & 27.897 & 27.337 & \textbf{27.535} & 26.089 & 27.471 & 26.589 & 26.119 \\ \cline{2-9}
                   & Peppers     & 29.986 & 29.239 & 29.202 & 26.992 & \textbf{29.356} & 27.732 & 27.069 \\ \hline\hline
\multirow{8}{*}{$\sigma=50$} & Airplane   & 26.193 & 25.407 & 25.408 & 24.811 & \textbf{25.471} & 25.3815  & 24.463 \\ \cline{2-9}
                   & Boat        & 25.537 & 24.789 & \textbf{24.871} & 24.472 & 24.825 & 24.802 & 23.980 \\ \cline{2-9}
                   & Couple      & 25.268 & 24.571 & \textbf{24.644} & 24.284 & 24.595 & 24.590 & 23.803 \\ \cline{2-9}
                   & Fingerprint & 22.662 & 22.383 & 22.361 & 21.871 & \textbf{22.442} & 22.367 & 21.419 \\ \cline{2-9}
                   & Hill        & 26.349 & 25.572 & \textbf{25.615} & 25.432 & 25.587 & 25.570 & 24.962 \\ \cline{2-9}
                   & Lena        & 27.476 & 26.622 & 26.519 & 26.243 & \textbf{26.699} & 26.526 & 25.811 \\ \cline{2-9}
                   & Man         & 25.710 & 25.075 & 25.098 & 24.782 & \textbf{25.128} & 25.053 & 24.316 \\ \cline{2-9}
                   & Peppers     & 27.235 & 26.411 & 26.236 & 26.052 & \textbf{26.518} & 26.291 & 25.524 \\ \hline\hline
\multirow{8}{*}{$\sigma=70$} & Airplane   & 24.550 & 23.684 & 23.481 & 23.068 & \textbf{23.967} & 23.772  & 22.464 \\ \cline{2-9}
                   & Boat        & 24.144 & 23.228 & 23.120 & 22.945 & \textbf{23.406} & 23.180 & 22.221 \\ \cline{2-9}
                   & Couple      & 23.907 & 23.079 & 22.945 & 22.844 & \textbf{23.279} & 23.053 & 22.036 \\ \cline{2-9}
                   & Fingerprint & 21.411 & 20.858 & 20.686 & 20.245 & \textbf{21.050} & 20.839 & 19.419 \\ \cline{2-9}
                   & Hill        & 25.023 & 24.054 & 23.898 & 24.056 & \textbf{24.315} & 24.106 & 23.169 \\ \cline{2-9}
                   & Lena        & 26.011 & 24.699 & 24.420 & 24.573 & \textbf{25.122} & 24.929 & 23.626 \\ \cline{2-9}
                   & Man         & 24.366 & 23.459 & 23.299 & 23.331 & \textbf{23.712} & 23.499 & 22.457 \\ \cline{2-9}
                   & Peppers     & 25.360 & 24.277 & 23.942 & 24.155 & \textbf{24.729} & 24.584 & 23.132 \\ \hline\hline
\multirow{8}{*}{$\sigma=100$} & Airplane  & 23.086 & 22.015 & 22.193 & 21.723 & \textbf{22.318} & 21.764 & 21.022 \\ \cline{2-9}
                   & Boat        & 22.737 & 21.690 & 21.842 & 21.623 & \textbf{21.853} & 21.446 & 21.024 \\ \cline{2-9}
                   & Couple      & 22.631 & 21.530 & \textbf{21.691} & 21.607 & 21.681 & 21.187 & 20.841 \\ \cline{2-9}
                   & Fingerprint & 20.079 & 19.253 & 19.384 & 18.991 & \textbf{19.417} & 18.687 & 17.982 \\ \cline{2-9}
                   & Hill        & 23.597 & 22.406 & 22.605 & 22.583 & \textbf{22.609} & 22.051 & 21.795 \\ \cline{2-9}
                   & Lena        & 24.531 & 22.660 & 22.881 & 22.749 & \textbf{23.051} & 22.430 & 21.650  \\ \cline{2-9}
                   & Man         & 23.016 & 21.832 & 22.017 & 21.981 & \textbf{22.067} & 21.531 & 21.051 \\ \cline{2-9}
                   & Peppers     & 23.648 & 22.258 & 22.447 & 22.364 & \textbf{22.708} & 22.085 & 21.144 \\ \hline
\end{tabular}

\end{table*}
\textbf{Comparison~$2$.} In this test, for each regularizer we fix $\lambda$ as the value achieving the best performance in above test. More specifically, we fix $\tau\eta=5$ for WNN, $\tau\eta=2$ for LNN, $\tau\eta=0.03$ for TNN and nuclear norm, $\tau\eta=8$ for $S_p$ with $p=0.1$, $\tau\eta=0.5$ for $S_p$ with $p=0.5$. Besides, we follow ~\cite{DBLP:journals/ijcv/GuXMZFZ17} and conduct WNN iteratively. The experimental results are used to serve as reference. The noise level is controlled by varying the parameter $\sigma$ in the set $\mathcal{S}_\sigma=\{10, 30, 50,70 ,100\}$. The PSNR
results are reported in Table~\ref{tab:res}. The information delivered by Table~\ref{tab:res} can be summarized as follows:
\begin{itemize}
  \item WNN-I achieves the best result in all cases due to conducting the reconstruction process iteratively. But, it provides only a small advantage over others that conduct only one iteration, especially the best one highlighted in bold.
  \item Non-convex relaxations outperform nuclear norm in most cases. Particularly,  the difference between NN and TNN is very small when $\sigma$ takes a small value.  But, residual non-convex relaxations  provide a large advantage over both TNN and NN. 
  \item The regularizer $S_p$ with $p=0.1$ achieves the best results in most cases. The reason is that the gap between $S_p$ with $p=0.1$ and nuclear norm is very small. A natural idea is: using the regularizer $S_p$ with $p<0.1$ can generate a better result. In practice, however, we do not support to use such a regularizer, because it treats all singular values almost equally. For instance, $10^{0.01}\thickapprox 1.0233$ and $1000^{0.01}\thickapprox 1.0715$.
\end{itemize}
\section{Conclusion and Discussion}\label{sec:6}
In this paper, we provided a comprehensive survey for low rank regularization including relaxations to rank-norm, optimization methods and practical applications. Differing from previous investigations that focus on nuclear norm, we pay more attention to the non-convex relaxations and corresponding optimization methods. Although the nuclear norm with solid theoretical guarantee has been widely used to solve various problems, the solution returned by it may deviate from the original problem significantly. Indeed, such a deviation can be alleviated by using non-convex relaxations. In order to promote the application of non-convex relaxations in solving practical issues, we give a detailed summarization for non-convex relaxations used in previous studies. Although the theoretical research over non-convex relaxation is still limited, numerous studies have experimentally verified that it often provide a large advantage over nuclear norm. 

An inevitable thing in solving LRR models~(without considering matrix factorization) is conducting SVD, which is time consuming for large scale data. So, introducing the techniques over approximate SVD is an important direction in the future~\cite{DBLP:journals/corr/abs-1708-00146}.
Besides, more recent efforts have been made in tensor learning~\cite{chen2013generalized,li2016simultaneous,zhang2015low,yokota2018missing,ko2018fast,zare2018extension,yokota2017simultaneous,lu2018tensor,zhou2018tensorlrr,lu2018exact}. But most of them are based on the convex relaxations. The problem of tensor completion based on non-convex regularization is considered in~\cite{ji2017non,yao2019efficient}. Generalizing tensor learning with non-convex relaxations to deal with practical issues is also promising in future.

\section*{ACKNOWLEDGMENTS}
This work was supported in part by the National Key Research and Development Program of China under Grant 2018YFB1403501, in part by the National Natural Science Foundation of China under Grant 61936014, Grant 61772427 and Grant 61751202, and in part by the Fundamental Research Funds for the Central Universities under Grant G2019KY0501.


\bibliographystyle{elsarticle-num}

\bibliography{LowRank}

\end{document}